\pdfoutput=1

\documentclass[11pt]{article}

\usepackage[final]{acl}

\usepackage{times}
\usepackage{latexsym}

\usepackage[T1]{fontenc}

\usepackage[utf8]{inputenc}

\usepackage{microtype}

\usepackage{inconsolata}

\usepackage{caption}
\usepackage{subcaption}
\usepackage{amsmath}
\usepackage{amssymb}
\usepackage{amsfonts}
\usepackage{xfrac}
\usepackage{bbm}
\usepackage{amsthm}
\usepackage{graphicx}
\usepackage{combelow}
\usepackage{qtree}
\usepackage{enumerate}
\usepackage{CJKutf8}
\usepackage{subcaption}
\usepackage{array}
\usepackage{multirow}
\usepackage{xcolor}
\usepackage{footmisc}

\usepackage{booktabs}
\usepackage{cleveref}
\Crefname{section}{\S}{\S\S}
\crefname{table}{Table}{}
\crefname{figure}{Figure}{}
\crefname{algorithm}{Algorithm}{}
\crefname{equation}{eq.}{eqs.}
\crefname{appendix}{App.}{}
\crefname{prop}{Prop.}{}
\crefformat{section}{\S#2#1#3}  

\usepackage[disable]{todonotes}

\makeatother

\newcommand{\VP}{\mathrm{VP}}
\newcommand{\Ss}{\mathrm{S}}
\newcommand{\NP}{\mathrm{NP}}
\newcommand{\PP}{\mathrm{PP}}

\newcommand{\Comp}{\mathrm{Comp}}
\newcommand{\Rel}{\mathrm{Rel}}

\newcommand{\Case}{\mathrm{Case}}

\newcommand{\ppls}{\mathboldface{\mathrm{PPL}}_\theta}
\newcommand{\ppl}{\mathrm{PPL}_\theta}

\usepackage[framemethod=TikZ]{mdframed}
\newcommand{\graybox}[1]{\begin{mdframed}[
    backgroundcolor=black!5,
    topline=false, bottomline=false, rightline=false, leftline=false,
    innertopmargin=0.5em,
    innerleftmargin=0.5em,
    innerbottommargin=0.5em,
    innerrightmargin=0.5em,
]{#1}
\end{mdframed}
}

\usepackage{url}

\newcommand{\mathboldface}[1]{\boldsymbol{#1}}
\newcommand{\bm}[1]{\mathboldface{#1}}
\newcommand{\rightbranch}{\includegraphics[width=0.9em]{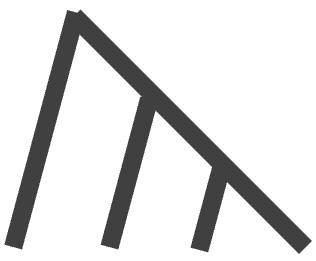}}
\newcommand{\leftbranch}{\includegraphics[width=0.9em]{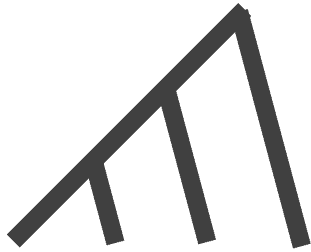}}
\newcommand{\mixedbranch}{\includegraphics[width=0.85em]{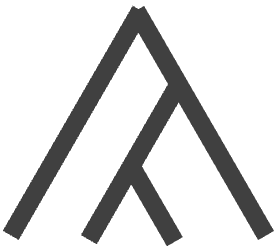}}

\newcommand{\orangecircle}{\includegraphics[width=0.7em]{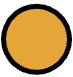}}
\newcommand{\bluesquare}{\includegraphics[width=0.7em]{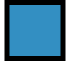}}
\newcommand{\greentriangle}{\includegraphics[width=0.8em]{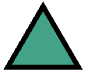}}
\newcommand{\joystick}{\includegraphics[width=0.8em]{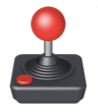}}
\newcommand{\Lstick}{\includegraphics[width=1em]{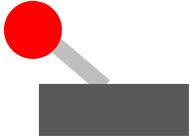}}
\newcommand{\Rstick}{\includegraphics[width=1em]{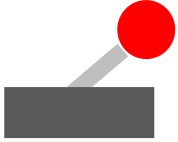}}

\newcommand{\hidden}[1]{}

\title{Emergent Word Order Universals \\ from Cognitively-Motivated Language Models}

\author{Tatsuki Kuribayashi$^{\rightbranch}$ \, Ryo Ueda$^{\leftbranch}$ \, Ryo Yoshida$^{\leftbranch}$ \, Yohei Oseki$^{\leftbranch}$ \\ \textbf{Ted Briscoe}$^{\rightbranch}$ \, \textbf{Timothy Baldwin}${}^{\rightbranch,\ \mixedbranch}$\\
        ${}^{\rightbranch}$Mohamed bin Zayed University of Artificial Intelligence \\
        ${}^{\leftbranch}$The University of Tokyo  \,
        ${}^{\mixedbranch}$The University of Melbourne \\
  \texttt{\{tatsuki.kuribayashi,ted.briscoe,timothy.baldwin\}@mbzuai.ac.ae} \\
  \texttt{\{ueda-ryo796,yoshiryo0617,oseki\}@g.ecc.u-tokyo.ac.jp}}

\begin{document}
\maketitle

\begin{abstract}
The world's languages exhibit certain so-called typological or implicational universals; for example, Subject-Object-Verb (SOV) languages typically use postpositions. Explaining the source of such biases is a key goal of linguistics.
We study word-order universals through a computational simulation with language models (LMs).
Our experiments show that typologically-typical word orders tend to have lower perplexity estimated by LMs with cognitively plausible biases: syntactic biases, specific parsing strategies, and memory limitations. 
This suggests that the interplay of cognitive biases and predictability (perplexity) can explain many aspects of word-order universals.
It also showcases the advantage of cognitively-motivated LMs, typically employed in cognitive modeling, in the simulation of language universals.
\newline
\newline
\vspace{0.1em}
\hspace{.5em}\includegraphics[width=1.1em,height=1.1em]{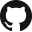}\hspace{.5em}\parbox{\dimexpr\linewidth-2\fboxsep-2\fboxrule}
  {\small \url{https://github.com/kuribayashi4/word-order-universals-cogLM}}
\end{abstract}

\section{Introduction}
\label{sec:intro}

There are thousands of attested languages, but they exhibit certain universal tendencies in their design. For example, Subject-Object-Verb (SOV) word order often combines with postpositions, while SVO order typically employs prepositions~\cite{greenberg1963some}. 
Researchers have argued that such implicational universals are not arbitrary but shaped by their advantage for humans~\cite{hawkins2004efficiency,Culbertson2012-bo,Culbertson2019-qo}.

Such language universals have been recently studied through neural-based computational simulation to elucidate the mechanisms behind the universals~\cite{Lian2023-cc}.
The languages which emerge, however, have typically not been human-like~\cite{chaabouni2019anti,chaabouni2019word,rita2022emergent,ueda2022word}. 
Such mismatch arguably stems from the lack of human-like cognitive biases in neural agents~\cite{galke2022emergent}, but injecting cognitive biases into systems and showing their benefits has proved challenging~\cite{Lian2021-ss}.

\begin{figure}[t]
    \centering
    \includegraphics[width=7.5cm]{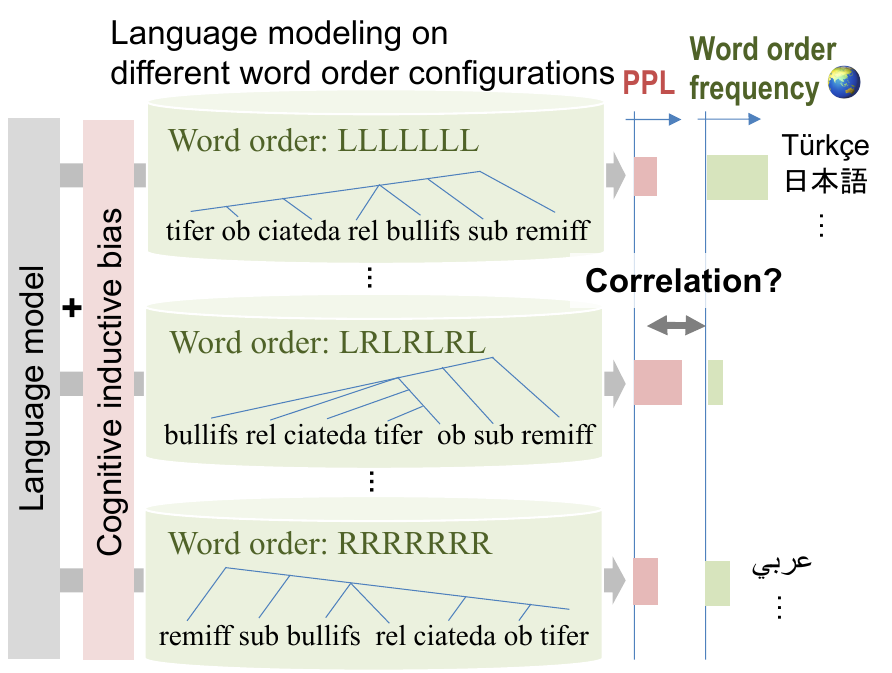}
    \caption{We compare the word orders that are challenging for LMs to those that are infrequent in attested languages (\cref{sec:design}).
    We examine the advantage of cognitively-motivated LMs (\cref{sec:model}) in simulating the word-order universals (the world's word-order distribution) with their inductive biases (\cref{sec:experiment}).}
    \label{fig:fig1}
\end{figure}

In this study, expanding on a study of word-order biases in language models (LMs:~\citet{white-cotterell-2021-examining}), we demonstrate the advantage of \textit{cognitively-motivated} LMs, which can simulate human cognitive load during sentence processing well~\cite{Hale2018FindingSearch,Futrell2020Lossy-ContextProcessing,Yoshida2021-ic,kuribayashi-etal-2022-context}, and thus predict many implicational word-order universals in terms of their inductive biases (Figure~\ref{fig:fig1}).
Specifically, we train various types of LMs in \textit{artificial languages} with different word-order configurations (\cref{sec:design}).
Our experiments show that perplexities estimated by cognitively-motivated LMs (\cref{sec:model}) correlate better with frequent word-order configurations in attested languages than standard LMs  (\cref{sec:experiment}).
This confirms that such biases are a potential source of the word-order universals as well as demonstrate the plausibility of cognitively-motivated LMs as models of human language processing.

\section{Related research}
\label{sec:related}

\subsection{Impossible languages and LMs}
\label{subsec:impossible_lang}
Generative linguistic theory has focused on delineating the impossible from possible languages in terms of universal grammar. \citet{Chomsky2023-bb} has recently argued that neural LMs cannot distinguish possible human languages from impossible, unnatural languages, based on the experiments by~\citet{mitchell-bowers-2020-priorless}, and are, therefore, of no interest to linguistic theory. \citet{Kallini2024-eq} challenge this claim, demonstrating that a standard transformer-based autoregressive model (GPT-2) assigns higher perplexity and greater surprisal to a range of artificially-generated, unattested, and thus putatively impossible candidate languages when compared to English.
In this work, by contrast, \textbf{we explore the ability of a variety of neural LMs to distinguish typologically rare combinations of word orders from the common attested combinations} as predicted by Greenberg's implicational universals~\cite{greenberg1963some}.

\hidden{
Our attempt to distinguish typologically common and rare word orders with LMs brings into question the argument about the poor role of LMs in linguistic theorizing~\cite{mitchell-bowers-2020-priorless,Chomsky2023-bb}.
They argue that LMs can not identify the distinctiveness of human language since the learning capability of LMs is too unlimited to distinguish between human language and impossible language. 
This study will contribute to clarifying which types of LMs are exactly unlimited against the (almost) impossible languages violating word-order universals~\cite{greenberg1963some}, although \textit{impossibility} of language can be defined by various aspects beyond word order variations~\cite{Moro2016-ej,Kallini2024-eq}.}

\subsection{The Chomsky hierarchy and LMs}
\label{subsec:chomsky}
We test how easy it is to learn a specific artificial language (with a specific word-order configuration) for certain LMs.
Such exploration is related to the investigation of the capabilities of neural LMs to generate formal, artificial languages in a specific class of the Chomsky hierarchy, such as irreducibly context-free (e.g., the Dyck languages) or mildly context-sensitive (e.g., \texttt{a$^{n}$b$^{n}$c$^{n}$}) languages~\cite{Weiss2018-aq,suzgun2019memory,hewitt-etal-2020-rnns,deletang2022neural}. 
While this line of research can elucidate whether specific models (LSTMs, transformers, etc.) are capable in principle of expressing and generalizing such languages, and thus also generating their putative analogs in natural language, in this work \textbf{we focus on artificial languages which are more human language-like} in that they exhibit a range of attested construction types, a more realistic vocabulary, and are less marked in terms of features like average sentence length, at least compared to the formal languages adopted in this line of research (\cref{app:artificial}).

\subsection{Word order preferences of LMs}
\label{subsec:word_order_lm}
Researchers have asked \textit{what kind of language is hard to language-model}~\cite{cotterell-etal-2018-languages,mielke-etal-2019-kind}, motivated by concern over whether the current language-modeling paradigm is equally suitable for all languages.
However, experiments using only attested language corpora made it difficult to single out impactful factors since they differ from each other in multiple dimensions~\cite{cotterell-etal-2018-languages,mielke-etal-2019-kind}.
Thus, prior studies adopted the use of \textit{artificially controlled} language(-like) data as a lens to quantify the inductive bias of models~\cite{Wang2016-md,white-cotterell-2021-examining,Hopkins2022-kf}.
Specifically, \citet{white-cotterell-2021-examining} pointed out some differences between LM's word-order preferences and common attested word orders (\textit{typological markedness}).
We expand on their research by exploring \textbf{which models, including cognitively-motivated ones, exhibit preferences more aligned with common typological patterns} (Figure~\ref{fig:fig1}).

\subsection{Cognitively-motivated LMs}
\label{subsec:cognitively_lm}
Cognitively-motivated LMs is a reference to a particular class of language model, typically examined and preferred in the cognitive modeling field. 
In brief, in the cognitive modeling field, the next-word probability and human reading behavior are compared~\cite{beinborn2023cognitive}, based on surprisal theory~\cite{Levy2008Expectation-basedComprehension,Smith2013TheLogarithmic}, to reverse-engineer what humans compute during language processing. Cognitively-motivated LMs typically yield a better fit to human behaviors than standard LMs. 
In Section~\ref{subsec:implementation_cognitively_lm}, we specifically focused on three aspects which are well established in the cognitive modeling literature: (i) syntactic bias~\cite{Hale2018FindingSearch}, (ii) parsing strategy~\cite{Resnik1992-zq,Yoshida2021-ic}, and (iii) memory limitations~\cite{Futrell2020Lossy-ContextProcessing,kuribayashi-etal-2022-context}.
Our interest is that \textbf{if language design is shaped by cognitive biases, such cognitively-motivated LMs should simulate the typological distribution of natural languages}.

\section{Preliminary}
\label{sec:design}
We explain the assumptions behind this research.

\subsection{Need for computational simulation}
\label{subsec:simulation}
This research aims to determine whether/how attested word order universals can stem from cognitive biases underlying human language processing.
Perhaps the strongest evidence would be obtained through counterfactual ablations.
That is, researchers should compare word orders that emerged in different, controlled scenarios: one emerged in our world with humans, and the other one emerged in the same world except for having non-human agents.
If the attested word order universals can emerge only in the scenario with humans, this can be evidence that the attested word order universal is related to human-like cognitive biases.
Nevertheless, unfortunately, such a controlled ablation is not feasible in the real world. 
Instead, this study adopts computational simulation to explore the source of word order universals.
That is, we virtually simulate several scenarios, each involving neural agents with different inductive biases, including human-like ones, and investigate under which biases our world's word order universals can emerge.

\subsection{Likning hypothesis}
\label{subsec:linking}

We further explain the assumptions underlying our computational simulations of the emergence of word order universals. 

\paragraph{Assumption 1:}
Given the theory that language has evolved to promote its processing efficiency~\cite{hawkins2004efficiency,Gibson2019-oe}, we posit that word orders frequently adopted in the world should be easily learned for human-like agents, and then, they can process it efficiently. 
That is, we assume that the frequency of word order $o$ within the attested languages should correlate with its (negative) learning/processing cost:
\begin{equation}
    \mathrm{frequency}(o) \propto -\mathrm{cost}(o) \;\;\mathrm{.}
    \label{eq:freq}
\end{equation}
\vspace{-0.5cm}
\graybox{See~\cref{subsec:wals} for word order configurations $o$.\\
See~\cref{subsec:freq} for the $\mathrm{frequency}(o)$ term. \\
See~\cref{subsec:artificial_data} and assumption 2 for the $\mathrm{cost}(o)$ term.
See~\cref{subsec:correl} for the correlation metrics $\propto$.
}

\paragraph{Assumption 2:}
We further posit that humans continuously predict the upcoming word during language processing, based on expectation-based theory~\cite{Levy2008Expectation-basedComprehension,Smith2013TheLogarithmic}.
That is, we assume that the $\text{cost}_\theta(o)$ for a particular model $\theta$ is determined by how the word order $o$ is difficult to learn and then incurs worse predictability through next-word prediction. 
Such a processing difficulty is measured as the average processing cost required to process sentences with word order $o$ for a particular model $\theta$ trained over a certain period of time of language acquisition (language modeling).
This can be quantified by \textit{perplexity} (PPL),\footnote{Using the average surprisal $-\frac{1}{|\bm x^o|}\sum_i \log p(x^o_i|\bm x^o_{<i})$ (entropy) instead of PPL is more aligned with surprisal theory~\cite{Smith2013TheLogarithmic}, but such a logarithmic conversion did not alter our findings (\cref{sec:analyses}).
Thus, we use PPL, following~\citet{white-cotterell-2021-examining}.} the geometric mean of word predictability, of a corpus with tokens $\bm x^o$ following word order $o$ under a learner $\theta$:
\begin{align}
    \mathrm{cost}_\theta(o) 
    &\sim  \underset{x^o_i\in \bm x^o}{\prod} p_\theta(x^o_i|\bm x^o_{<i})^{-\frac{1}{|\bm x^o|}} \;\;\mathrm{,} \\
    &:= \ppl(\bm x^o) \;\;\mathrm{.}
\end{align}
\noindent
Here,  $p_\theta(x^o_i|\bm x^p_{<i})$ is a probability of the $i$-th word in context for a model $\theta$ that is trained on a corpus with word order $o$.
We analyze \textbf{which LM $\theta$ computes $\ppl(\bm w^o)$ that well correlates with its typological frequency} as expected in Eq.\ref{eq:freq}.
\graybox{
See~\cref{sec:model} for LMs $\theta$ we examine. \\
See~\cref{subsec:results} and~\cref{subsec:regression} for main results. \\
See~\cref{subsec:linking_function} for other variants of PPLs. \\
See~\cref{subsec:parseability} for the connection to the bi-dimensional view of communicative efficiency.
}

\section{Problem settings}
\subsection{Word-order configurations}
\label{subsec:wals}

\textit{Branching directionality} $\joystick \in\{\texttt{L},\texttt{R}\}$, the concept of whether a dependent phrase is positioned left (\texttt{L})  or right (\texttt{R}) of its head in a particular constituent, is a key component of the typological theory.
We define six word-order parameters to classify attested languages as shown in Table~\ref{tbl:switches}.
For example, the parameter $\joystick^{\boldsymbol{\Ss}}$ determines the order of the subject $\boldsymbol{\NP}$ and the $\boldsymbol{\VP}$.
Then, the combination of word order parameter assignments defines a word order configuration $o=(\joystick^{\boldsymbol{\Ss}},\joystick^{\boldsymbol{\VP}},\joystick^{\boldsymbol{\PP}},\joystick^{\boldsymbol{\NP}},\joystick^{\boldsymbol{\Rel}},\joystick^{\boldsymbol{\Case}}) \in \mathcal{O}:=\{\texttt{L},\texttt{R}\}^6$, denoted by a sequence of \texttt{L}/\texttt{R}.
For example, \texttt{LRLLLR} $\in \mathcal{O}$ is the configuration where all phrases, except for $\boldsymbol{\VP}$ and $\boldsymbol{\Case}$, are left-branching.
Such parameter combinations result in $2^6=64$ word order configurations.

\subsection{Frequencies of word order}
\label{subsec:freq}
\begin{figure}[t]
    \centering
    \includegraphics[width=\linewidth]{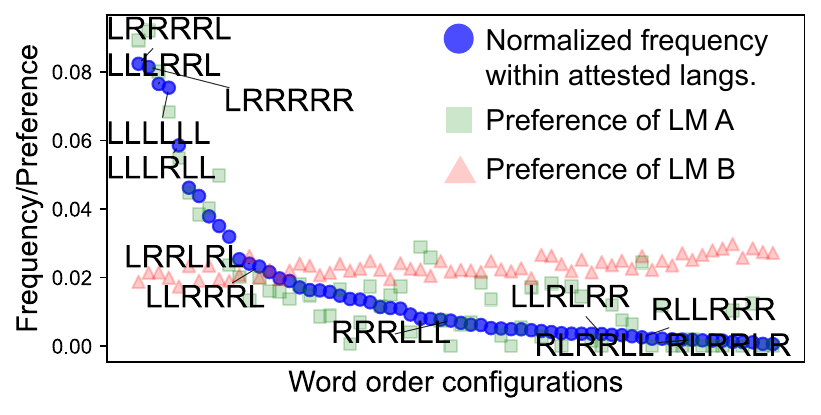}
    \caption{The frequency distribution of $2^6=64$ word-order configurations within attested languages (blue points) sorted in descending order. Suppose particular LMs \texttt{A}/\texttt{B} prefer word order as green/red points. The LM \texttt{A} (green points) is considered to have typologically more aligned inductive bias than the LM \texttt{B} (red points).}
    \label{fig:fig2}
\end{figure}

The 64 word-order configurations are not uniformly distributed among attested languages (blue points in Figure~\ref{fig:fig2}).
This distribution is estimated by the frequency of word orders in The World Atlas of Language Structures (WALS:~\citet{wals}),\footnote{We used the word order statistics of 1,616 languages, out of 2,679, where at least one parameter is annotated. If a particular parameter is missing or non-binary (\texttt{X}), one count is distributed between its compatible word orderings, e.g., \texttt{L\underline{L}LL\underline{L}R}, \texttt{L\underline{L}LL\underline{R}R}, \texttt{L\underline{R}LL\underline{L}R}, and \texttt{L\underline{R}LL\underline{R}R} each gets a 1/4 count for \texttt{L\underline{X}LL\underline{X}R}. See~\cref{app:wals} for the details of the WALS.} which is also denoted as a vector $\bm f=[\mathrm{freq}(\texttt{LLLLLL}), \mathrm{freq}(\texttt{LLLLLR}),..., \mathrm{freq}(\texttt{RRRRRR})]$.
Notably, particular configurations, typically with harmonic (consistent) branching-directionality, e.g., \texttt{LLLLLL}, \texttt{LRRRRR}, are common; such a skewed distribution  (\textit{typological markedness} or \textit{word-order universals}) has been studied from multiple perspectives typically tied with human cognitive biases~\cite{vennemann1974analogy,gibson2000dependency,Briscoe2000-gg,levy2005probabilistic,Christiansen2008-lj,Culbertson2012-bo,Temperley2018-bg,Futrell2019-zh,Futrell2020-rm}.

\begin{table}[t]
\centering
\fontsize{9}{9.5}\selectfont
\setlength{\tabcolsep}{1pt}

\begin{tabular}{lll}
\toprule
\textbf{Param.}        & \multicolumn{1}{c}{\texttt{L} \Lstick} & \multicolumn{1}{c}{\texttt{R} \Rstick}  \\
\cmidrule(r){1-1} \cmidrule(r){2-2} \cmidrule(r){3-3}  
$\joystick^{\boldsymbol \Ss}$ & {\color{red}Cat} {\color{cyan}eats}.  &  {\color{cyan}Eats} {\color{red}cat}.  \\
$\joystick^{\boldsymbol \VP}$ &  Cat {\color{red}mouse} {\color{cyan}eats}. & Cat {\color{cyan}eats} {\color{red}mouse}. \\
$\joystick^{\boldsymbol \PP}$  & {\color{red}Cat} {\color{cyan}table}  {\color{orange}on} eats.  & {\color{red}Cat} {\color{orange}on} {\color{cyan}table} eats.    \\
$\joystick^{\boldsymbol \NP}$  & {\color{cyan}Small} {\color{red}cat} eats. & {\color{red}Cat} {\color{cyan}small} eats. \\
$\joystick^{\boldsymbol \Rel}$  &  {\color{cyan}Likes milk} that {\color{red}cat} eats. & {\color{red}Cat} that {\color{cyan}likes milk} eats. \\
$\joystick^{\boldsymbol \Case}$   & {\color{red}Cat}{\color{cyan}\texttt{-sub}} eats. & {\color{cyan}\texttt{Sub-}}{\color{red}cat} eats.  \\
\bottomrule
\end{tabular}

\caption{Word-order parameters and example constructions with different assignments, \texttt{L} or \texttt{R} (See Apps.~\ref{app:artificial} and~\ref{app:wals} and~\citet{white-cotterell-2021-examining} for details).}
\label{tbl:switches}
\end{table}

\subsection{Processing costs of word order}
\label{subsec:artificial_data}
\paragraph{Artificial languages:}
We quantify which word orders are harder for a particular LM.
Here, we adopt\footnote{We introduce the $\boldsymbol{\Case}$ parameter determining the position of case marker, while \citet{white-cotterell-2021-examining} fixed it to be \texttt{L}.
We omitted the $\boldsymbol{\Comp}$ switch controlling the complementizer position, e.g., ``that,'' due to the lack of large-scale statistics on its order. We experimented with prepositional and postpositional complementizer settings in each of the 64 settings and used the average perplexities of the two settings.} the set of artificial languages created by~\citet{white-cotterell-2021-examining} as a lens to quantify the LMs' biases.
These languages share the same default probabilistic context-free grammar (PCFG) and differ from each other only in their word-order configuration $o \in \mathcal{O}$ (\cref{subsec:wals}) overriding the word order rules in the default grammar, resulting in $2^6=64$ corpora with different word order $o$.
Note that the 64 corpora generated have the same probabilities under the respective grammar and gold parser; thus, differences in language-modeling difficulties can only stem from the model's biases.
See~\cref{app:artificial} for the detailed configurations of artificial languages.

\paragraph{Quantifying perplexities:}
We train an LM on each corpus with word order $o$ and measure the PPL of tokens $\bm x^o$ in the respective held-out set.
Repeatedly conducting the training/evaluation across the 64 corpora produces a PPL score vector, $\ppls=[\ppl(\bm x^{\texttt{LLLLLL}}), \ppl(\bm x^{\texttt{LLLLLR}}),\ppl(\bm x^{\texttt{LLLLRL}}), \\ ..., \ppl(\bm x^{\texttt{RRRRRR}})]$, which indicates the \textit{word-order preferences} of an LM.

\subsection{Frequency--perplexity correlations}
\label{subsec:correl}
\paragraph{Global correlation:}
We measure the Pearson correlation coefficient $r(\cdot,\cdot)$ between word order frequencies $\bm f$ (\cref{subsec:freq}) and their \textit{negative} $\ppls$ (\cref{subsec:artificial_data}), considering \textit{lower} PPL is better.
We call $r(\bm f, -\ppls)$ \textit{global correlation}.
A high global correlation indicates that the LMs' word-order preferences reflect typological markedness.

\paragraph{Local correlation:}
\citet{white-cotterell-2021-examining} reported that simulating the word-order distribution among subject, object, and verb (SOV$\succ$SVO$\succ$VOS$\succ$OVS), which is determined by the first two parameters of $\joystick^{\boldsymbol \Ss}$ and $\joystick^{\boldsymbol \VP}$, is challenging.
Therefore, we assess how easy it is to simulate the markedness of the other parameters' assignments.
Specifically, we measure a relaxed version of the correlation ignoring the subject, object, and verb order (\textit{local correlation}), which is defined by the averaged correlation within each base word-order group: 
SOV ($\joystick^{\boldsymbol \Ss}=\texttt{L}$, $\joystick^{\boldsymbol \VP}=\texttt{L}$), SVO ($\joystick^{\boldsymbol \Ss}=\texttt{R}$, $\joystick^{\boldsymbol \VP}=\texttt{L}$), OVS ($\joystick^{\boldsymbol \Ss}=\texttt{L}$, $\joystick^{\boldsymbol \VP}=\texttt{R}$), and VOS ($\joystick^{\boldsymbol \Ss}=\texttt{R}$, $\joystick^{\boldsymbol \VP}=\texttt{R}$).

{\small
\begin{align}
\begin{split}
        & \frac{1}{4}(r(\bm f^{\texttt{SOV}}, -\ppls^{\texttt{SOV}}) + r(\bm f^{\texttt{SVO}}, -\ppls^{\texttt{SVO}}) \\
    &+ r(\bm f^{\texttt{OVS}}, -\ppls^{\texttt{OVS}}) + r(\bm f^{\texttt{VOS}}, -\ppls^{\texttt{VOS}})) \;\;\mathrm{.}
\end{split}
\end{align}
}

\noindent
Here, $\bm f^{\texttt{X}}$ and $\ppls^{\texttt{X}}$ are the list of frequencies and perplexities, limited to the languages with the assignments of $\joystick^{\boldsymbol \Ss}$ and $\joystick^{\boldsymbol \VP}$ corresponding to {\texttt{X}}.
If this relaxed correlation is high and the global correlation is low, the ordering of subject, object, and verb remains challenging, and indeed, this is the case (see \cref{subsec:sov_discussion}).

\section{Models}
\label{sec:model}
We examine 23 types of uni-directional LMs, which adopt subwords split by byte-pair-encoding~\cite{Sennrich2016-uu} and are trained for 10 epochs.
For each type of LM, we train/evaluate five models with different random seeds.
In each run, 20K sentences (train:dev:test is 8:1:1) are generated from PCFG to train/evaluate the model.
See~\cref{app:model} for model details.

\subsection{Standard LMs}
We test the PPL estimated by a Transformer \cite{vaswani2017}, LSTM~\cite{Hochreiter1997-rt}, simple recurrent neural network (SRN)~\cite{Elman1990-on}, and N-gram LMs.\footnote{Neural LMs are trained with the fairseq toolkit~\cite{Ott2019-xn}. N-gram LMs are trained with the KenLM toolkit~\cite{Heafield2011-ce} with Kneser-Ney smoothing.}
See~\cref{app:hyperparams} for the model details.

\subsection{Cognitively-motivated LMs}
\label{subsec:implementation_cognitively_lm}
We further test cognitively-motivated LMs employed in cognitive modeling and incremental parsing, mentioned in~\cref{subsec:cognitively_lm}.
We target three properties: (i) syntactic inductive/processing bias, (ii) parsing strategy, and (iii) working memory limitations, following recent works in cognitive modeling research~\cite{Dyer2016-nq,Hale2018FindingSearch,Resnik1992-zq,Oh2021-ln,Yoshida2021-ic,Futrell2020Lossy-ContextProcessing,kuribayashi-etal-2022-context}.

\begin{figure*}[t]
    \centering
    \includegraphics[width=15cm]{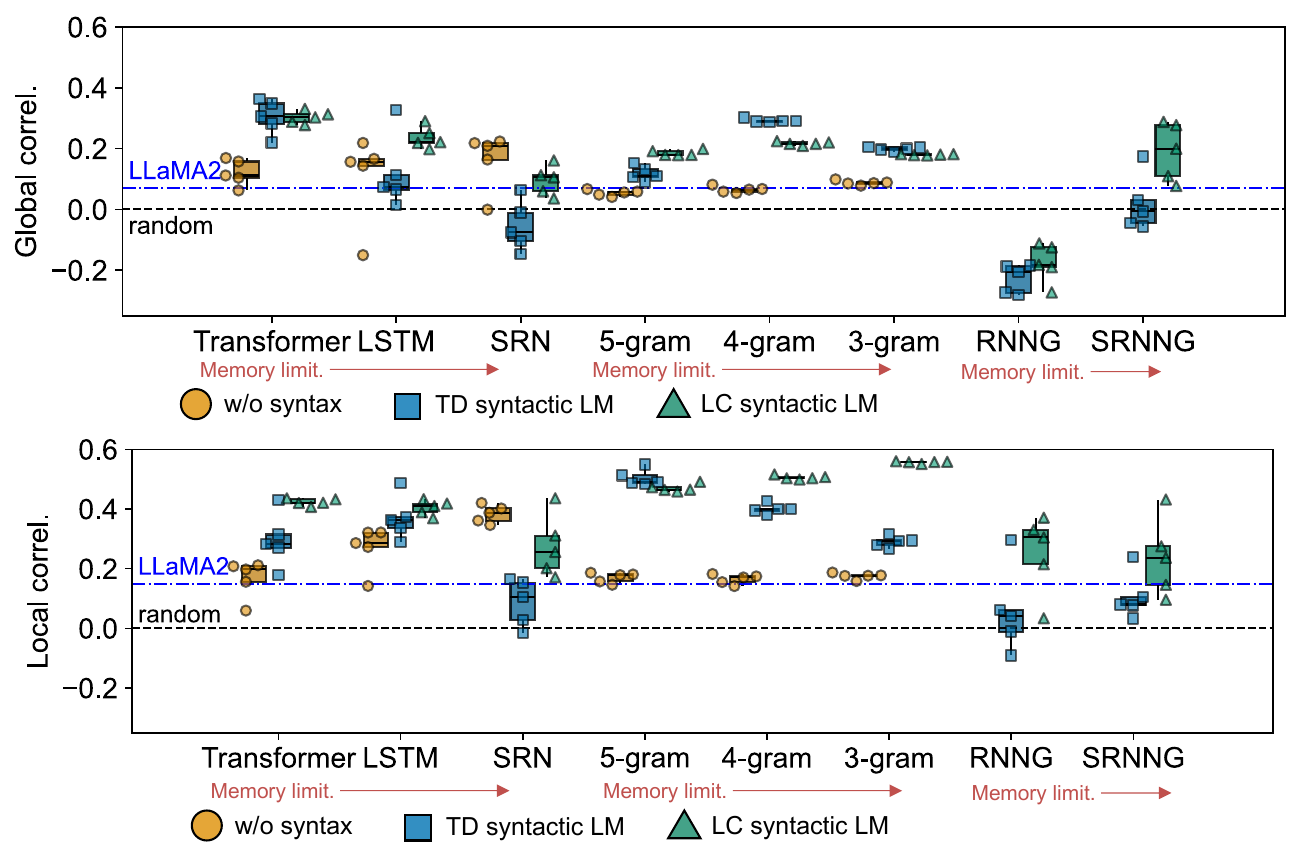}
\caption{The results of global/local correlations. Each point corresponds to each run. Their colors and shapes denote the syntactic bias of the models. The TD and LC variants in the Transformer, LSTM, SRN, and N-gram settings correspond to the respective PLMs. The box presents the lower/upper quartiles.}
\label{fig:correls}
\end{figure*}

\paragraph{Syntactic LMs and parsing strategy:}
We begin with \textit{syntactic} LMs to explore the cognitively-motivated LMs. 
Under a particular word order configuration $o$, they jointly predict tokens $\bm x^o$ and their syntactic structures $\bm y^o$ by incrementally predicting parsing actions $\bm a^o$, such as ``\texttt{NT($\Ss$) NT($\NP$) GEN(I) REDUCE($\NP$)...''}:
\begin{align}
    p_\theta(\bm x^o, \bm y^o) &= \prod_t p_\theta(a^o_t|\bm a^o_{<t}) \;\;\;\mathrm{.}
\end{align}
\noindent
Here, we examine two commonly-adopted strategies to convert the $(\bm x^o, \bm y^o)$ into the actions $\bm a^o$: top-down (TD) and left-corner (LC) strategies~\cite{kuncoro-etal-2017-recurrent}.\footnote{TD and LC are also denoted as pre-order and in-order traversals, respectively.} The LC strategy is theoretically expected to estimate more human-like cognitive loads than the TD~\cite{Abney1991-vl,Resnik1992-zq} in a sense that LC parsers can estimate the cognitive load of center-embedding constructions, typically with disharmonic word order configurations, which tend to be avoided by humans~\cite{miller1964free}.\footnote{We adopted the arc-standard LC strategy, following~\citet{Kuncoro2018-ws} and~\citet{Yoshida2021-ic}. Strictly speaking, an arc-eager LC strategy is cognitively plausible, and an arc-standard one has similar characteristics with bottom-up traversal~\cite{Resnik1992-zq}. That is, our LC results may be biased towards \texttt{L} assignments.\label{footnote:left-corner}}

\paragraph{Memory limitation:}
Humans generally have limited working memory~\cite{Miller1956-xz} and struggle with processing long-distance dependencies during sentence processing~\cite{hawkins1994performance,Gibson2019-oe,Hahn2020-sb}.
We thus expect that model architectures with more severe memory access, e.g., in the order of SRN$\succ$LSTM$\succ$Transformer, have such human-like biases and exhibit higher correlations with the word-order universals.
We also apply such a memory limitation to syntactic LMs (next paragraph).
We expect that memory-limited syntactic LMs can softly reflect the stack-depth, a standard concept of working memory demand~\cite{Abney1991-vl,noji2014left}, since the requirement of more stack depth needs the model to maintain longer parsing histories, presumably incurring a high cost for memory-limited syntactic LMs~\cite{jin-schuler-2020-memory}.

\paragraph{Implementations:} We train (i) Parsing-as-Language-Model (PLM)~\cite{Choe2016-ff} and (ii) recurrent neural network grammar (RNNG)~\cite{Dyer2016-nq,kuncoro-etal-2017-recurrent}, using the syntactic structures obtained during generating the corpus from PCFG (\cref{subsec:artificial_data}).
PLMs are the same as standard LMs expect that they are trained on the action sequences $\bm a$. 
Four PLMs with different architectures (Transformer, LSTM, SRN, and N-gram) are tarined.
RNNGs also predict the action sequences, but they have an explicit composition function to compute phrase representations.
We use the stack-only RNNG implementation by~\citet{Noji2021-mv}, and we newly introduce its memory-limited version (simple recurrent neural network grammar; SRNNG), where (Bi)LSTMs are replaced by SRNs.
Note that the cognitive plausibility of RNNGs has been reported in both cognitive modeling~\cite{Hale2018-gd,Yoshida2021-ic} and language generalization test~\cite{Kuncoro2018-ws,wilcox-etal-2019-structural}.
Henceforth, \textit{syntactic LMs} refer to the PLMs and (S)RNNGs.

\paragraph{PPL computation:} We measure the PPL over action sequences in each word order $o$ when quantifying the word-order preference of syntactic LMs (\cref{subsec:artificial_data}): $\ppl(\bm x^o, \bm y^o) := \prod_t p_\theta(a^o_t|\bm a^o_{<t})^{\frac{1}{|\bm a^o|}}$.
We also examine a token-level predictability $\ppl(\bm x^o)$ in \cref{subsec:parseability} and \cref{app:beam}, but such variations did not alter the conclusions.

\subsection{Baselines}
We set two baselines: (i) a chance rate with random assignments of perplexities (gray lines in Figure~\ref{fig:correls}), and (ii) perplexities estimated by pre-trained LLaMA2 (7B)~\cite{Touvron2023-dc}, a representative of the large language models (LLMs), prompted with several example sentences (blue lines in Figure~\ref{fig:correls}) (\cref{app:llama}) as a naive baseline.

\section{Experiments}
\label{sec:experiment}
We compare the LM's word-order preferences with attested word-order distributions (\cref{sec:experiment}).
Then, we further analyze what kind of word-order combinations LMs prefer (\cref{sec:analyses}).

\subsection{Results}
\label{subsec:results}
Figure~\ref{fig:correls} shows global and local correlations (see \cref{app:results} for the full results).
The TD \bluesquare\ and LC \greentriangle\ variations of the Transformer, LSTM, SRN, and N-gram LMs correspond to the PLMs with their respective architecture.
We expect syntactic LMs with the LC strategy to exhibit higher correlations than the LMs without syntactic biases (\greentriangle$>$\orangecircle) and those with cognitively unmotivated TD syntactic bias (\greentriangle$>$\bluesquare).

\paragraph{Most LMs beat the chance rate:}
Overall, most global and local correlations were higher than the random baseline, reproducing the general trend that common word orders induce lower PPL~\cite{Hahn2020-dq}.\footnote{With a one-sample, one-sided t-test, models except for LSTM LM, TD SRN PLM, TD RNNG, LC RNNG, TD SRNNG yielded global correlations significantly larger than zero, and models except for TD RNNG yielded local correlations significantly larger than zero.}
As a sanity check, we also observed that the LLaMA-2 exhibited weaker correlations than cognitively-motivated LMs; the current success of LLMs is orthogonal to our results.

\paragraph{Syntactic biases and parsing strategies:}
First, the LC syntactic LMs  generally outperformed the standard LMs (\greentriangle$>$\orangecircle) in each setting except for SRNs.
This indicates the \textbf{advantage of cognitively-motivated syntactic biases in simulating the word-order universals.}
Second, LMs with the LC strategy tend to exhibit higher correlations than TD syntactic LMs  (\greentriangle$>$\bluesquare), especially in terms of local correlation.
That is, \textbf{the cognitively-motivated LC parsing strategy better simulates the word-order implicational universals.}
Note that RNNGs, on average, exhibited low correlations, although they are often claimed to be cognitively plausible LMs.

\paragraph{Memory limitation:}
The results show that memory-limited models tend to exhibit better correlations, with the exception of PLMs. 
In particular, \textbf{RNNGs typically benefited from memory limitations} (SRNNG$\succ$RNNG), while PLMs did not (SRN$\prec$Transformer).
This implies a superiority of RNNGs' memory decay over hierarchical tree encoding to PLMs' simple linear memory decay. 

\subsection{Statistical test}
\label{subsec:regression}
\paragraph{Settings:}
We statistically test the advantage of cognitively-motivated factors toward higher correlations.
Specifically, we train the following regression model to predict the global or local correlation scores obtained in the experiment (\cref{subsec:results}):\footnote{We used the \texttt{statsmodels}~\cite{seabold2010statsmodels}.}
\begin{align}
\begin{split}
        &r(\bm f, -\ppls) \sim \mathrm{ModelClass}(\theta) \\ &\;\;\;\;+ \mathrm{MemLim}(\theta) + \mathrm{Syntax}(\theta) + \mathrm{LC}(\theta)  \;\;\mathrm{.}
\end{split}
\end{align} 
\noindent
Here, $\mathrm{ModelClass}$ denotes the coarse type (e.g., neural model or not) of the model $\theta$ yielding the respective correlation score, $\mathrm{MemLim}$ denotes its strength of context limitation (higher is severer, e.g., SRN$\succ$LSTM$\succ$Transformer), $\mathrm{Syntax}$ denotes whether the model is syntactic LMs (1 for syntactic LMs; otherwise 0), and $\mathrm{LC}$ denotes whether the model uses the LC strategy (1 for LC syntactic LMs; otherwise 0).
Positive coefficients for these features indicate their contribution to higher correlations.
See~\cref{app:regression} for the details of the regression.

\paragraph{Results:} We observe that the coefficients for the $\mathrm{Syntax}$ and $\mathrm{LC}$ features were significantly larger than zero with one-sample, two-sided t-test in both cases of predicting global and local correlations.\footnote{$p=0.07$ for the $\mathrm{Syntax}$ and $p<0.05$ for the $\mathrm{LC}$ in the case of global correlation. $p<0.01$ for the $\mathrm{Syntax}$ and $p<0.01$ for the $\mathrm{LC}$ in the case of local correlation.}
The coefficient for the $\mathrm{MemLim}$ feature was not significantly larger than zero when targeting all the models ($p>0.1$); however, when PLMs were excluded, the coefficient of the $\mathrm{MemLim}$ feature was also significantly larger than zero with one-sample, two-sided t-test ($p<0.001$ in both global and local correlations) as suggested in~\cref{subsec:results}.
To sum up, \textbf{these corroborate the findings in~\cref{subsec:results}.}

\begin{figure}[t]
    \centering
    \includegraphics[width=\linewidth]{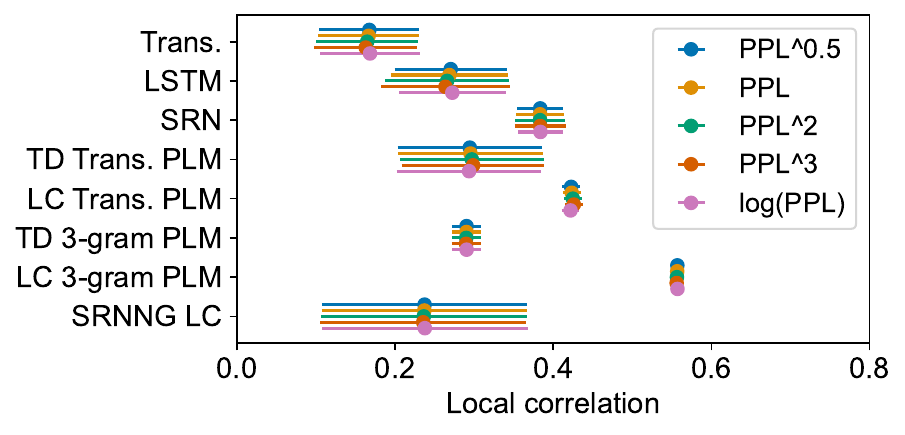}
    \caption{Mean and standard deviation of local correlations with different linking functions: PPL of order $k$ and logarithmic PPL
    }
    \label{fig:linking_hypothesis}
\end{figure}

\subsection{Linking functions}
\label{subsec:linking_function}
The experiments so far have assumed the linearity between PPL and word-order frequency (Eq.~\ref{eq:freq})---did this choice bias our results?
We investigated various linking functions between PPL and word order frequency: the perplexity of order $k$ and logarithmic PPL (entropy).
Our findings robustly hold with other linking functions (full results are in Appendix~\ref{app:link}).
Figure~\ref{fig:linking_hypothesis} illustrates LMs' local correlations under differently converted perplexities.

\section{Analyses: what kind of word order is particularly (un)preferred?}
\label{sec:analyses}

\subsection{\{S,O,V\} word-order biases}
\label{subsec:sov_discussion}
\begin{figure}[t]
    \centering
    \includegraphics[width=\linewidth]{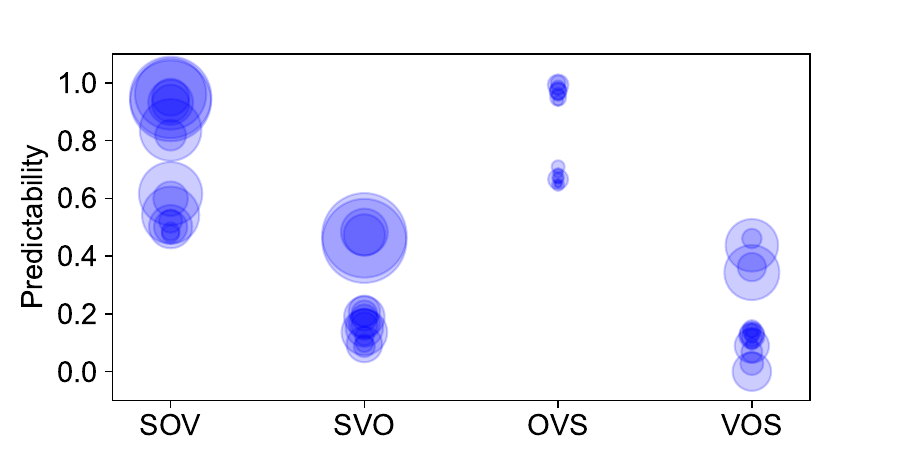}
    \caption{Illustration of the relationship between predictability (y-axis) and word order frequency in each of the four base-order groups (SOV, SVO, OVS, and VOS).
    Each circle corresponds to each word order; larger ones are frequent word orders.
    Predictability is the negative PPL converted through the min-max normalization; thus higher predictability indicates lower PPL. The results are from the 3-gram PLM with the LC strategy.
    }
    \label{fig:intra_svo}
\end{figure}

\paragraph{Observation:}
\citet{white-cotterell-2021-examining} reported that LMs could not show the subject, object, verb word-order biases attested in natural language  (SOV$\succ$SVO$\succ$VOS$\succ$OVS).
Even our cognitively-motivated LMs did not overcome this limitation, based on the global correlations being consistently lower than the local ones (\cref{subsec:results}; Figure~\ref{fig:correls}).
This tendency is visualized in Figure~\ref{fig:intra_svo}; within each base group (SOV, SVO, OVS, VOS), common word orders tend to obtain high predictability (i.e., lower PPL; bigger circles are at the top) except OVS-order's high predictability and SVO-order's low predictability.
This made it clear that predictability generally explains word-order universals, but the markedness of word orders among subject, object, and verb must stem from additional factors.

\paragraph{Implication:}
This finding is consistent with: (i) humans arguably have an agent-first bias in event cognition, which could be the source of the subject-initial word order~\cite{doi:10.1126/sciadv.abn8464}; and (ii) agent-preference in human sentence processing cannot be explained by surprisal estimated by neural LMs in N400 amplitudes modeling~\cite{10.1162/nol_a_00121}.
Our findings corroborate that cognitively-motivated LMs still lack such a human-like bias.
Orthogonally, the artificial language ignores some important linguistic aspects, such as information structure~\cite{gundel1988universals,verhoeven2015thematic,ranjan-etal-2022-discourse}, which may explain subject-object order; refining the artificial data would also be one direction to explore in future work.

\subsection{Branching directionality preferences}

\begin{figure*}[t]
    \centering
    \includegraphics[width=15cm]{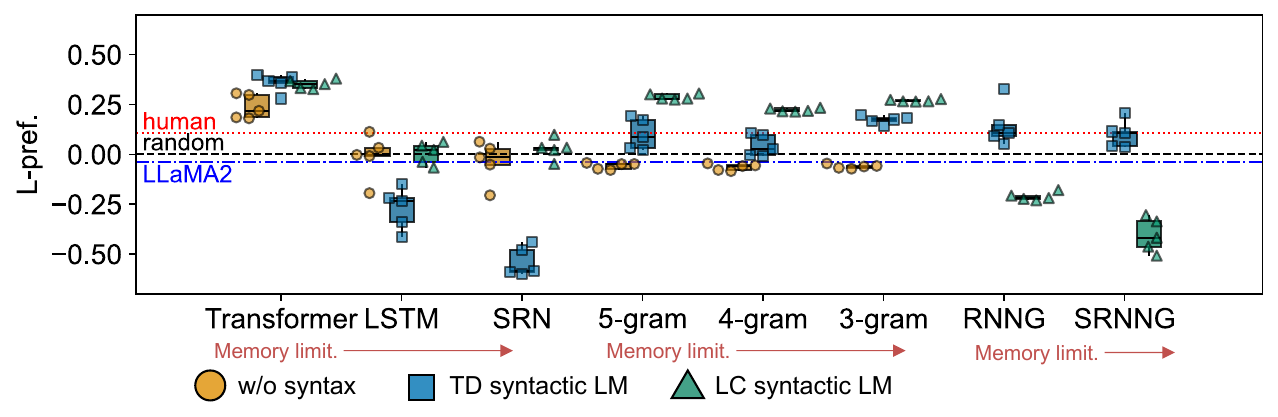}
\caption{The results of branching directionality scores. Each point corresponds to each run. The colors and shapes denote the syntactic bias of the models. The TD and LC variants in the Transformer, LSTM, SRN, and N-gram settings correspond to the respective PLMs. The box presents the lower/upper quartiles.}
\label{fig:harmonic_lpref}
\end{figure*}

\paragraph{Settings:}
Human languages, on average, do not favor either left- or right-branching~\cite{wals}.
 Given this, we measure how strongly a model prefers a specific branching directionality.
We calculate the Pearson correlation (\texttt{L}-pref.) between negative PPL and the number of \texttt{L} assignments $\mathrm{\#}\texttt{L}(\cdot)$\footnote{For example, $\mathrm{\#}\texttt{L}(\texttt{LLLRLL})=5$.} of the word order:
\begin{align}
        &r(\bm l, -\ppls) \;\;\mathrm{,}\\
        &\bm l := [\mathrm{\#}\texttt{L}(\texttt{\small{LLLLLL}}), \mathrm{\#}\texttt{L}(\texttt{\small{LLLLLR}}),...,\mathrm{\#}\texttt{L}(\texttt{\small{RRRRRR}})] \;\mathrm{.}
\end{align}
\noindent
As a sanity check, the word-order frequency distribution of attested languages, indeed, is weakly correlated (0.11) with the left-branching directionality.
Thus, LMs are not expected to have an extremely high or low \texttt{L}-pref. score.
Notably, the branching bias of LMs/parsers has been of interest in the NLP research~\cite{li-etal-2020-branching,li-etal-2020-empirical,ishii-miyao-2023-tree}.

\paragraph{Results:}
Figure~\ref{fig:harmonic_lpref} shows the results of branching preferences.
LMs with the TD strategy are theoretically expected to have a lower \texttt{L}-pref. score (favoring right-branching) than the LC models~\cite{Resnik1992-zq} (\bluesquare$<$\greentriangle).
While the PLMs faithfully reflect such characteristics, RNNGs, surprisingly, exhibited opposite trends, suggesting the challenge in controlling the inductive bias of neural syntactic LMs.
We also observed architecture-dependent branching preference; Transformers prefer left-branching, while LSTMs prefer right-branching as suggested by~\citet{Hopkins2022-kf}. 
Such architecture-dependent biases seem to have more impact on the branching preferences than the parsing strategies in PLMs.

\section{Connection to predictability and parsability}
\label{subsec:parseability}

\paragraph{Background:}
Human language is arguably designed to minimize \textit{complexity} (how unpredictable symbols are) while maintaining \textit{informativity} (how easy it is to extract a message from symbols)~(\citet{I_Cancho2003-kd,Piantadosi2012-de,Kemp2012-pv,Frank2012-uu,Kirby2015-kz,Kanwal2017-qe,Gibson2019-oe,Xu2020-at,Hahn2020-dq}; \textit{inter alia}).
We revisit this bi-dimensional view.
Concretely, \citet{Hahn2020-dq} showed that both PPL of an LM and parsability for a (not-cognitively-motivated) parser~\cite{Kiperwasser2016-qf} explain word-order universals.

\paragraph{Oveview:}
We demonstrate that predictability (PPL)\footnote{Predictability is typically measured as entropy, but again, the choice of entropy or PPL did not substantially change the correlation scores (See~\cref{subsec:linking_function} and ~\cref{app:link}).} of syntactic LMs entails parsability.
That is, they can provide a more concise information-theoretic measurement of word-order universals (\textit{syntactically-biased predictability}).

\paragraph{Settings:}
We decompose the performance of syntactic LMs into two parts: token-level perplexity $\ppl(\bm x^o)$  (predictability) and parsing performance $\mathrm{parse}_\theta(\bm x^o, \bm y^o)$ (parsability), using word-synchronous beam-search~\cite{Stern2017-ef}.
When computing the token-level predictability $\ppl(\bm x^o)$, next-word probability is computed while predicting the upcoming partial syntactic trees.\footnote{$\ppl(\bm x^o) := \prod_t p_\text{stx}(x^o_t|\bm x^o_{<t})^{\frac{1}{|\bm x^o|}}$. $p_\text{stx}(x^o_t|\bm x^o_{<t}) := \sum_{y'\in \mathcal{Y}(\bm x^o_{<t})} p(x^o_t, y'| \bm x^o_{<t})$ Here, given a context $\bm x_{<t}$, a set of its upcoming compatible partial syntactic trees  $\mathcal{Y}(\bm x_{<t})$ is predicted. Next word $x^o_t$ is predicted by each candidate parse $y'\in \mathcal{Y}(\bm x^o_{<t})$, then such predictions are merged over $\mathcal{Y}(\bm x^o_{<t})$.}
We measure $\mathrm{parse}_\theta(\bm x^o, \bm y^o)$ as the F1-score of the top-1 parse found with the beam search.\footnote{Evalb (\url{https://nlp.cs.nyu.edu/evalb/}) was used.}
Then, we test whether the parsability factor contributes to explaining the frequency of word order $o$ in addition to PPL, using the following nested regression models:
\begin{align}
    \nonumber
    \mathrm{\textbf{Base:}}\ \mathrm{freq}(o) &\sim \ppl(\bm x^o) \;\;\mathrm{,} \\
    \nonumber
    \mathrm{\textbf{+Parse:}}\ \mathrm{freq}(o) &\sim \ppl(\bm x^o)+\mathrm{parse}_\theta(\bm x^o, \bm y^o) \;\mathrm{.}
\end{align} 

\paragraph{Results:}
The increase in log-likelihood scores of the $\mathrm{\textbf{+Parse}}$ model over the $\mathrm{\textbf{Base}}$ model is not significant with the likelihood-ratio test ($p>0.1$) in all the RNNG settings (\{TD, LC\}$\times$\{SRNNG, RNNG\}$\times$\{seeds\}).\footnote{We only tested RNNGs given the limited availability of batched beam-search implmentations~\cite{Noji2021-mv}.}
That is, at least in our setting, we cannot find an advantage of parsability over predictability in explaining word-order universals.
This may be because the next-word prediction for the syntactic LMs is explicitly conditioned by the parsing states, which might sufficiently bias the predictability measurements to reflect parsability.

\begin{figure}[t]
    \centering
    \includegraphics[width=.95\linewidth]{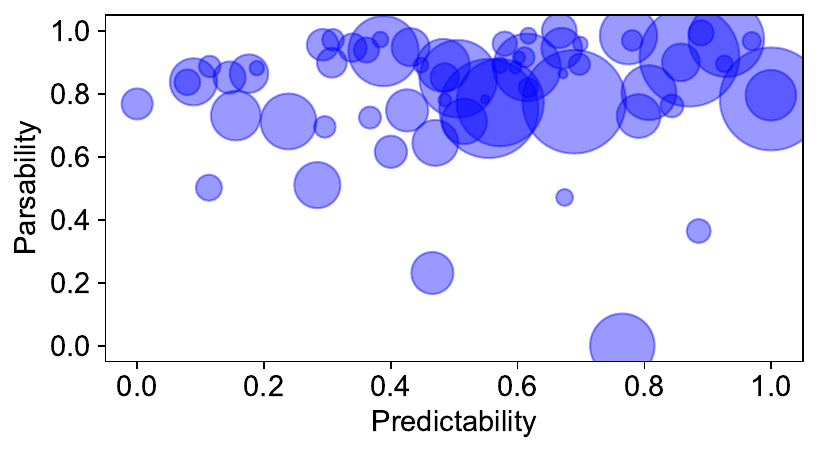}
    \caption{Predictability and parsability of each word order. These measurements are converted through the min-max normalization to be [0, 1] scale (higher is better). Each circle corresponds to each word order; larger ones are frequent word orders.}
    \label{fig:parsability}
\end{figure}

Figure~\ref{fig:parsability} also illustrates the predictability and parsability estimated by the LC SRNNG. 
Here, the predictability identifies more types of word orders as typologically marked (left small circles) than the parsability does (bottom small circles).
This is contrary to the picture of both predictability and parsability as complementary factors explaining word-order universals~\cite{Hahn2020-dq}.
Nevertheless, our artificial data could be easier to parse than natural language in the sense that our data have less structural ambiguity; thus, future work should explore this using more real artificial language.

\section{Conclusions}
We have shown that cognitively-motivated LMs better led to the emergence of word-order universals than standard LMs.
From the linguistic typology perspective, we provide computational evidence that the universals emerge from cognitive biases, which has been challenging to demonstrate in previous work~\cite{Lian2021-ss,galke2022emergent}.
From the cognitive modeling perspective, our results demonstrate that cognitively-motivated LMs have human-like biases that are sufficient to replicate some human word-order universals.
Our results also clarified what typological patterns can still not be modeled by cognitively-motivated LMs (agent-first bias), as well as opened the direction to relate LMs in the traditional linguistic theory of delineating possible language from the impossible.

\section*{Limitations}
\paragraph{Artificial language:}
We used artificial data to quantify the LMs' inductive/processing biases for word-order configurations.
While the use of artificial languages has typically been adopted to conduct controlled experiments (\cref{subsec:chomsky}), such artificial data lack some properties of natural languages, such as the semantic relationships between the linguistic constituents (\cref{subsec:sov_discussion}). 
In future work, we hope to devise further artificially controlled languages that exhibit some of these properties. 

In addition, our PCFG implementation of artificial languages and word-order configurations unreasonably limited the candidate word-order variations; for example, the production rules (i) $\boldsymbol{\Ss}\rightarrow \boldsymbol{\NP_\text{subject}}\ \boldsymbol{\VP}$, and (ii) $\boldsymbol{\VP}\rightarrow \boldsymbol{\NP_\text{object}}\ \boldsymbol{\mathrm{Verb}}$ could not produce variations with V and O apart (VSO and OSV) (Table~\ref{tbl:grammar}).
Exploring more flexible ways to create and cover artificial langauge variations will be another interesting direction.

Furthermore, our used data~\cite{white-cotterell-2021-examining} is relatively small scale, which might incur unintended bias in LM performance, although there is a perspective to analyze inductive bias via measuring training efficiency~\cite{kharitonov2021what,Warstadt2023-rp}.

\paragraph{Estimating word-order distribution:}
The word order frequency estimates derived from WALS might also be biased; for example, Indo-European languages tend to have richer meta-linguistic information in WALS, although our study takes the statistics from as many as 1,616 languages into account.
A richer estimation of missing parameter information is desirable. 
As a more general concern, the frequency of a word-order configuration can be estimated in various ways, such as the number of native speakers or the number of language families adopting a particular word order.
Furthermore, word order variation can be inherently non-binary~\cite{Levshina2021-cm}.
Our study, as an initial foray, relied on the number of unique languages, a commonly used metric in linguistic typology research~\cite{wals,hammarstrom2016linguistic}, considering that other approaches raise additional complications, such as an estimation of the speaker numbers or language family variability.

\section*{Ethical Considerations}
We only used artificial language, which does not have information with potential risks, e.g., human privacy data.
One concern is the bias in our word order frequency estimates; this might have led to biased conclusions, e.g.,  diminishing the impact of minority languages, although we used the largest publicly available data (WALS) to date.
We used AI assistance tools within the scope of ``Assistance purely with the language of the paper'' described in the ACL 2023 Policy on AI Writing Assistance.

\section*{Acknowledgements}
We are grateful to Ana Brassard, Benjamin Heinzerling, Tatsuro Inaba, Go Inoue, and Yova Kementchedjhieva for their insightful comments on an early version of this study.
This work was supported by JST PRESTO Grant Number JPMJPR21C2, Japan.

\bibliography{custom}

\clearpage
\appendix

\begin{table*}[t]
\centering
\fontsize{8}{9.5}\selectfont
\setlength{\tabcolsep}{2pt}

    \begin{tabular}{lccc}
    \toprule
    & Right-branching & Mixed-branching & Left-branching \\
    \cmidrule(r){1-1} \cmidrule(lr){2-2} \cmidrule(lr){3-3} \cmidrule(l){4-4}
    Parameteres & \texttt{RRRRRR} & \texttt{LRRRLR} & \texttt{LLLLLL} \\
    \cmidrule(r){1-1} \cmidrule(lr){2-2} \cmidrule(lr){3-3} \cmidrule(l){4-4}
    &
        {\tiny
        \Tree 
    [.{$\Ss$}  
        [.{$\VP$}  
            [.{$\mathrm{Verb}$} \textit{strovokicizeda} ] !\qsetw{1cm}
        ] !\qsetw{2cm}
            [.{$\NP$}
                [.{$\NP$} \textit{fusbenders} ] !\qsetw{1cm}
                [.{$\Rel$} \textit{rel} ] !\qsetw{0.5cm}
                [.{$\VP$}  
                    [.{$\mathrm{Verb}$} \textit{povify} ]  !\qsetw{1cm}
                        [.{$\NP$} 
                            [.{$\mathrm{Pronoun}$} \textit{me} ] 
                        ] !\qsetw{1cm}
                ] !\qsetw{1.5cm} 
            ] !\qsetw{2cm}
    ] 
    }
    &
    {\tiny
    \Tree 
    [.{$\Ss$}  
            [.{$\NP$}  
                [.{$\NP$} \textit{fusbenders} ] !\qsetw{0.5cm}
                [.{$\Rel$} \textit{rel} ] !\qsetw{1cm}
                [.{$\VP$} 
                    [.{$\mathrm{Verb}$} \textit{povify} ] 
                        [.{$\NP$}  
                            [.{$\mathrm{Pronoun}$} \textit{me} ] !\qsetw{0.5cm}
                        ] 
                ] !\qsetw{1.1cm}
        ] 
        [.{$\VP$} 
            [.{$\mathrm{Verb}$} \textit{strovokicizeda} ] 
        ] !\qsetw{0.5cm}
    ]
    }
    &
        {\tiny
     \Tree 
    [.{$\Ss$} 
            [.{$\NP$}  
                [.{$\VP$}  
                        [.{$\NP$} 
                            [.{$\mathrm{Pronoun}$} \textit{me} ] 
                        ] !\qsetw{1cm}
                    [.{$\mathrm{Verb}$} \textit{povify} ] 
                ]  
                [.{$\Rel$} \textit{rel} ] !\qsetw{0.5cm}
                [.{$\NP$} \textit{fusbenders} ] !\qsetw{1cm}
            ] !\qsetw{1.5cm}
        [.{$\VP$}  
            [.{$\mathrm{Verb}$} \textit{strovokicizeda} ]
        ]  !\qsetw{2cm}
    ]
    }
    \\
    \bottomrule
    \end{tabular}

    \caption{Example sentences and their structures generated with different word-order configurations.}
    \label{tbl:trees}
\end{table*}

\section{Details of artificial languages}
\label{app:artificial}

Table~\ref{tbl:grammar}\footnote{The corresponding Table is positioned in the later part of the Appendix for readability.\label{note2}} shows the default grammar to create artificial language data, which adopts the configuration of~\citet{white-cotterell-2021-examining}.
The ``relevant parameter'' column indicates the parameter \joystick\ that controls the order of right-hand items in the respective production rule (the order is swapped when the parameter assignment is \texttt{R}).
Note that the subcategory of non-terminal symbols (e.g., $\VP\_\mathrm{S}$) was only used for generating the data; in the final data for training/evaluating syntactic LMs, these subcategories are omitted (e.g., $\VP\_\mathrm{S}$ should be $\VP$), and the resulting uninformative edge in the syntactic structure (e.g., $\VP \rightarrow \VP$) was also removed.
Table~\ref{tbl:trees} shows the example of a sentence with different word-order configurations.
Different word order parameters yield a syntactic structure with different branching directionalities; for example, the constituency tree of \texttt{LLLLLL} extends to the bottom left (left-branching).
The average sentence length was 11.8 tokens, and the average tree depth was 9.1.
The vocabulary size of pseudowords is 1,314 same as~\citet{white-cotterell-2021-examining}.

\begin{table}[t]
    \centering
\fontsize{8}{9}\selectfont
\setlength{\tabcolsep}{2pt}
\begin{tabular}{lp{4cm}}
\toprule
All languages in WALS & 2,679 \\
Targeted languages & 1,616 \\
Targeted parameters & 9,696 (=1,616$\times$6) \\
Missing parameters & 3,343 \\
\cmidrule(r){1-1} \cmidrule(lr){2-2}
\texttt{LLXXXX} (SOV) & 46.7\% \\
\texttt{LRXXXX} (SVO) & 34.3\% \\
\texttt{RLXXXX} (VOS) & 3.6\% \\
\texttt{RRXXXX} (OVS) & 15.5\% \\
\cmidrule(r){1-1} \cmidrule(lr){2-2}
$s^{\boldsymbol \Ss}$ ($\boldsymbol{\Ss} \rightarrow \boldsymbol{\NP}_\textbf{subj}\ \boldsymbol{\VP}$) & 82A Order of Subject and Verb~\cite{wals-82} \\
$s^{\boldsymbol \VP}$ ($\boldsymbol{\VP} \rightarrow \boldsymbol{\NP}_\textbf{obj}\ \textbf{Verb}$) & 83A Order of Object and Verb~\cite{wals-83} \\
$s^{\boldsymbol \PP}$ ($\boldsymbol{\PP} \rightarrow \textbf{Prep}\ \boldsymbol{\NP}$) & 85A Order of Adposition and Noun Phrase~\cite{wals-85} \\
$s^{\boldsymbol \NP}$ ($\boldsymbol{\NP} \rightarrow \textbf{Adj}\ \boldsymbol{\NP}$) & 87A Order of Adjective and Noun~\cite{wals-87} \\
$s^{\boldsymbol \Rel}$ ($\boldsymbol{\NP} \rightarrow \boldsymbol{\VP}\ \textbf{Rel}\ \boldsymbol{\NP}$) & 90A Order of Relative Clause and Noun~\cite{wals-90} \\
$s^{\boldsymbol \Case}$ ($\boldsymbol{\NP} \rightarrow \boldsymbol{\NP}\ \textbf{Case}$)  & 51A Position of Case Affixes~\cite{wals-51} \\
\bottomrule
\end{tabular}

\caption{Statistics of the WALS data and the source of word-order configuration information}
\label{tbl:wals}
\end{table}

\section{WALS data statistics}
\label{app:wals}
Table~\ref{tbl:wals} shows the statistics of the WALS data and the details about word order parameters.
Out of the 2,679 languages listed in the WALS, 1,616 languages are involved, and approximately $2/3$ of their word order parameters were annotated in the WALS; the missing values are completed as explained in~\cref{subsec:wals} (footnote 1).
As a sanity check, we observe the ratio of the assignments of the first two parameters ($\joystick^{\boldsymbol \Ss}$ and $\joystick^{\boldsymbol{\VP}}$), which controls the order of subject, object, and verb; these approximately replicate the ratio reported in~\citet{wals-81}, e.g., SOV and SVO orders occupy over 80\% of languages.

\begin{table}[t]
\centering
\fontsize{8}{9}\selectfont
\setlength{\tabcolsep}{2pt}
\begin{tabular}{p{5cm}p{2cm}}
\toprule
Data/model  &  Licence \\
\cmidrule(r){1-1} \cmidrule(r){2-2}
Artificial data~\cite{white-cotterell-2021-examining} & MIT \\
WALS~\cite{wals} & Creative Commons CC-BY 4.0 \\
Fairseq~\cite{Ott2019-xn} & MIT \\
RNNG~\cite{Noji2021-mv} & MIT \\
KenLM~\cite{Heafield2011-ce} & LGPL \\
LLaMA-2~\cite{Touvron2023-dc} &  LLAMA 2 Community License \\
Sentencepiece~\cite{Kudo2018-xb} & Apache 2.0 \\
\bottomrule
\end{tabular}
\caption{Licence of the data and models}
\label{tbl:licence}
\end{table}

\section{Model details}
\label{app:model}
The license of the used models/data is listed in Table~\ref{tbl:licence}; all of them are used under their intended use.
All the models were trained/tested with a single NVIDIA A100 GPU (40GB).
All of the experiments were done within approximately 600 GPU hours.
The LLaMA-2 (7B) model was used via the hugging face toolkit~\cite{wolf-etal-2020-transformers}.

\subsection{Parsing strategies}
\begin{figure*}[t]
    \centering
    \includegraphics[width=15cm]{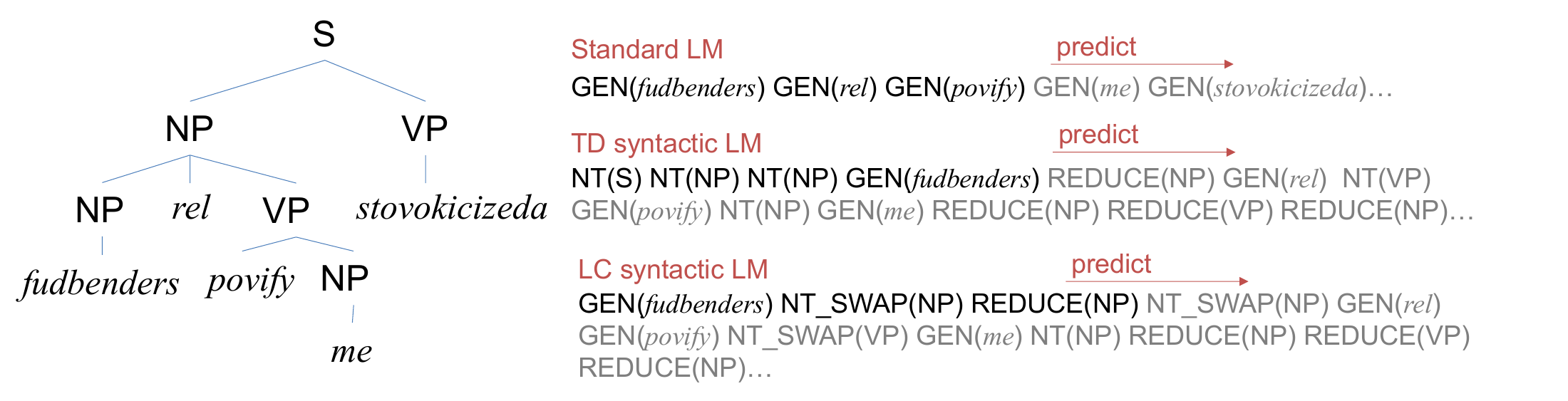}
    \caption{Different parsing strategy converts a syntactic structure into a different parsing action sequence. PLMs and RNNGs predict such action sequences with different model architectures. }
    \label{fig:traversals}
\end{figure*}

Figure~\ref{fig:traversals} shows the parsing actions converted with different strategies (TD and LC).
PLMs are trained to predict such a sequence of parsing actions in a left-to-right manner.

\subsection{Hyperparamteres}
\label{app:hyperparams}

Tables~\ref{tbl:hyper_params} and~\ref{tbl:hyper_params_rnngs} show the hyperparameters of LMs,\footref{note2} which basically follow their default settings.
Standard LMs and PLMs use the same hyperparameters.
Their vocabulary size is set to $512$.

\subsection{Word order preference of LLaMA-2}
\label{app:llama}
We employed few-shot settings instead of full-finetning with the limits in computational cost.
Specifically, we create a prompt consisting of the instruction ``\textit{The below sentences are written in an artificially created new language:}'' and ten example sentences extracted from the respective training set. 
The probability of each test sentence is computed conditioned with this prompt, and aggregating these probabilities results in the PPL of an entire corpus.

\section{Results}
\label{app:results}
The full results of the experiment (\cref{sec:experiment}) and analysis (\cref{sec:analyses}) are shown in Table~\ref{tbl:results}.
This also shows the top-3 preferred word orders by each LM, which demonstrates the model-dependent differences in their word-order preferences.
We also include the baseline of average stack depth required to process sentences for each parsing algorithm in each word order as a standard measurement of memory cost.

\subsection{Beam-search in RNNG/SRNNGs}
\label{app:beam}
Table~\ref{tbl:beam} shows the results of RNNG/SRNNGs with and without word-synchronous beam search~\cite{Stern2017-ef}.
The settings without beam-search are adopted in~\cref{sec:experiment}, and the advantages of memory limitation (SRRNG) and the LC strategies were replicated even with beam-search, where the token-level perplexity $\mathrm{PPL}(\bm x)$ is used (\cref{subsec:parseability}).

\subsection{Results with different linking functions}
\label{app:link}

Tables~\ref{tbl:results_k0.5}, \ref{tbl:results_k2}, \ref{tbl:results_k3}, and \ref{tbl:results_klog} show the detailed results with different linking functions ($\mathrm{PPL}^{1/2}$, $\mathrm{PPL}^{2}$, $\mathrm{PPL}^{3}$, $\log \mathrm{PPL}$) between model-computed complexity  measurements and word order frequencies.
Experiments with different linking hypotheses did not alter the conclusions. This supports the generality of our findings.

\begin{table}[t]
    \centering
    \fontsize{8}{8.5}\selectfont
    \setlength{\tabcolsep}{1pt}
    \begin{tabular}{lp{1.8cm}p{1.2cm}p{1cm}p{1cm}}
    \toprule
    Model & $\mathrm{ModelClass}$ (categorical) & $\mathrm{MemLim}$ (int) & $\mathrm{Syntax}$ (binary) & $\mathrm{LC}$\ \ \ \ \ \ \ (binary) \\
    \midrule
    Transformer  & NLM & 0 & False & False \\
    LSTM  & NLM & 1 & False  & False \\
    SRN   & NLM & 2 & False & False \\
    \cmidrule(r){1-5}
    Trans. PLM TD & NLM & 0 & True & False \\
    Trans. PLM LC & NLM & 0 & True & True \\
    LSTM PLM  TD & NLM & 1 & True & False \\
    LSTM PLM  LC & NLM & 1 & True & True \\
    SRN PLM  TD & NLM & 2 & True & False \\
    SRN PLM  LC & NLM & 2 & True & True \\
    \cmidrule(r){1-5}
    Word 5-gram  & CLM & 0 & False & False \\
    Word 4-gram  & CLM & 1 & False & False \\
    Word 3-gram  & CLM & 2 & False & False \\
    \cmidrule(r){1-5}
    5-gram PLM TD & CLM & 0 & True & False \\
    5-gram PLM LC & CLM & 0 & True & True \\
    4-gram PLM TD & CLM & 1 & True & False \\
    4-gram PLM LC & CLM & 1 & True & True \\
    3-gram PLM TD & CLM & 2 & True & False \\
    3-gram PLM LC & CLM & 2 & True & True \\
    \cmidrule(r){1-5}
    RNNG  & RNNG & 0 & True & False \\
    RNNG LC & RNNG & 0 & True & True \\
    SRNNG  & RNNG & 1 & True & False \\
    SRNNG LC & RNNG & 1 & True & True \\
    \bottomrule
    \end{tabular}
        \caption{Features used for the regression analysis}
        \label{tbl:regression}
    \end{table}

\section{Details of regression analysis}
\label{app:regression}

We explored which factor impacts the global/local correlation scores obtained by various LMs $\theta$.
As explained in~\cref{subsec:regression}, we train a regression model to predict the correlation score obtained by a particular LM $\theta$, given the features characterizing the LMs: 
\begin{align}
\begin{split}
    \mathrm{Correl}_\theta \sim&\ \mathrm{ModelClass}(\theta) + \mathrm{MemLim}(\theta) \\
    &+ \mathrm{Syntax}(\theta) + \mathrm{LC}(\theta)  \;\;\mathrm{.}
\end{split}
\end{align} 
\noindent
Table~\ref{tbl:regression} shows the features used for the regression analysis in~\cref{subsec:regression}.
The regression model is trained with the ordinary least squares setting, using \texttt{statsmodel} package in Python~\cite{seabold2010statsmodels}.

\begin{table*}[t]
    \centering
    \fontsize{9}{8.5}\selectfont
    \setlength{\tabcolsep}{2pt}
    
\begin{tabular}{llcrrrc}
    \toprule
    Model & Lim. & Stx. &  \multicolumn{1}{c}{Global $r$}$\uparrow$ & \multicolumn{1}{c}{Local $r$}$\uparrow$ & \texttt{L}-pref. $\rightarrow$ & Top3 langs. \\
\cmidrule(r){1-1} \cmidrule(r){2-2} \cmidrule(r){3-3}  \cmidrule(lr){4-4} \cmidrule(l){5-5} \cmidrule(l){6-6} \cmidrule(l){7-7}
Natural Lang. & & & 100.0 & 100.0 & 10.5 & \texttt{LRRRRL}, \texttt{LRRRRR}, \texttt{LLLRRL} \\
\cmidrule(r){1-3} \cmidrule(l){4-7}
Transformer & & & 12.1 $\pm$ 4.3 & 16.7 $\pm$ 6.4 & 23.8 $\pm$ 6.1 & \texttt{LLLLLL}, \texttt{LLRLLL}, \texttt{RLRLLL} \\
LSTM & \textcolor{gray}{\checkmark} & & 10.7 $\pm$ 14.7 & 26.9 $\pm$ 7.4 & -1.2 $\pm$ 11.3 & \texttt{RLRLLL}, \texttt{RLLLLL}, \texttt{RRRRRR} \\
SRN  & \checkmark & & 16.3 $\pm$ 9.4 & 38.3 $\pm$ 3.0 & -3.6 $\pm$ 10.4 & \texttt{RLLLLL}, \texttt{RLRLLL}, \texttt{RLRRLL} \\
\cmidrule(r){1-3} \cmidrule(l){4-7}
Word 5-gram & \checkmark & & 5.4 $\pm$ 1.0 & 17.0 $\pm$ 1.7 & -5.8 $\pm$ 1.6 & \texttt{RRLRRR}, \texttt{RRRRRR}, \texttt{LRLRRR} \\
Word 4-gram & \checkmark & & 6.5 $\pm$ 1.0 & 16.5 $\pm$ 1.6 & -6.4 $\pm$ 1.6 & \texttt{RRLRRR}, \texttt{RRRRRR}, \texttt{LRLRRR} \\
Word 3-gram & \checkmark & & 8.8 $\pm$ 0.7 & 17.5 $\pm$ 1.0 & -6.1 $\pm$ 1.0 & \texttt{RRRRRR}, \texttt{RRLRRR}, \texttt{LRLRRR} \\
\cmidrule(r){1-3} \cmidrule(l){4-7}
Trans. PLM & & TD & 30.4 $\pm$ 5.7 & 29.5 $\pm$ 9.1 & 35.9 $\pm$ 4.7 & \texttt{LLRRLL}, \texttt{LLRRRL}, \texttt{LLRLLL} \\
Trans. PLM & & LC & 30.3 $\pm$ 2.1 & 42.3 $\pm$ 1.1 & 35.2 $\pm$ 2.3 & \texttt{LLLLLL}, \texttt{LLRLLL}, \texttt{LLRRLL} \\
LSTM PLM & \textcolor{gray}{\checkmark} & TD & 11.9 $\pm$ 12.2 & 37.0 $\pm$ 7.3 & -27.1 $\pm$ 10.5 & \texttt{LRRRRL}, \texttt{LRLRRL}, \texttt{LRRRRR} \\
LSTM PLM & \textcolor{gray}{\checkmark} & LC & 23.6 $\pm$ 3.6 & 40.4 $\pm$ 2.5 & 0.5 $\pm$ 5.5 & \texttt{RLLRRR}, \texttt{LLLLLL}, \texttt{LLLRLL} \\
SRN PLM & \checkmark & TD & -5.4 $\pm$ 8.3 & 8.7 $\pm$ 7.8 & -53.7 $\pm$ 7.4 & \texttt{RLRRRR}, \texttt{LRLRRR}, \texttt{RLLRRR} \\
SRN PLM & \checkmark & LC & 9.5 $\pm$ 4.9 & 27.5 $\pm$ 10.5 & 2.8 $\pm$ 5.2 & \texttt{RLRRLL}, \texttt{RLRRRR}, \texttt{RLLRLL} \\
\cmidrule(r){1-3} \cmidrule(l){4-7}
5-gram PLM & \checkmark & TD & 11.8 $\pm$ 2.4 & 50.4 $\pm$ 2.8 & 10.2 $\pm$ 7.8 & \texttt{RLLRRL}, \texttt{RLRRRL}, \texttt{RLLLLL} \\
5-gram PLM & \checkmark & LC & 18.6 $\pm$ 0.9 & 47.0 $\pm$ 1.3 & 28.8 $\pm$ 1.4 & \texttt{LLLRLL}, \texttt{RLLRLL}, \texttt{RLRRLL} \\
4-gram PLM & \checkmark & TD & 29.2 $\pm$ 0.6 & 40.0 $\pm$ 1.6 & 4.4 $\pm$ 5.4 & \texttt{LLRRRL}, \texttt{RLRRRL}, \texttt{LLRRRR} \\
4-gram PLM & \checkmark & LC & 21.7 $\pm$ 0.6 & 50.6 $\pm$ 0.6 & 22.2 $\pm$ 0.9 & \texttt{RLLRLL}, \texttt{RLRRLL}, \texttt{LLLRLL} \\
3-gram PLM & \checkmark & TD & 19.9 $\pm$ 0.7 & 29.0 $\pm$ 1.8 & 17.3 $\pm$ 2.0 & \texttt{RLLRRL}, \texttt{LLLRRL}, \texttt{RLLRRR} \\
3-gram PLM & \checkmark & LC & 18.0 $\pm$ 0.2 & 55.7 $\pm$ 0.3 & 27.0 $\pm$ 0.5 & \texttt{RLLRRR}, \texttt{RLLRRL}, \texttt{RLRRRR} \\
\cmidrule(r){1-3} \cmidrule(l){4-7}
RNNG & & TD & -22.6 $\pm$ 4.7 & 6.0 $\pm$ 14.5 & 14.5 $\pm$ 10.8 & \texttt{RLLRLL}, \texttt{RRRRLL}, \texttt{RRRLLL} \\
RNNG & & LC & -17.6 $\pm$ 6.4 & 25.1 $\pm$ 13.4 & -21.2 $\pm$ 2.0 & \texttt{RRRRRL}, \texttt{RRLLRL}, \texttt{RRLRRL} \\
SRNNG & \checkmark & TD & 2.0 $\pm$ 9.3 & 10.7 $\pm$ 7.8 & 10.2 $\pm$ 7.0 & \texttt{RLLRRR}, \texttt{RLRRRR}, \texttt{RLLRRL} \\
SRNNG & \checkmark & LC & 19.0 $\pm$ 9.6 & 23.7 $\pm$ 13.0 & -40.6 $\pm$ 8.5 & \texttt{LRRRRR}, \texttt{LRLRRR}, \texttt{LLLRRR} \\
\cmidrule(r){1-3} \cmidrule(l){4-7}
LLaMA2 (7B) & &  & 6.9 $\pm$ 31.0 & 15.4 $\pm$ 2.5 & -4.6 $\pm$ 31.0 & \texttt{LRLLLL}, \texttt{LRRLLL}, \texttt{LRLRLL} \\
\cmidrule(r){1-3} \cmidrule(l){4-7}
Stack depth & & TD & -47.5 $\pm$ 0.2 & -12.0 $\pm$ 0.6 & -56.2 $\pm$ 1.3 & \texttt{RRLRRR}, \texttt{RRLLRR}, \texttt{RRRRRR} \\
Stack depth & & LC & -13.3 $\pm$ 0.3 & -4.8 $\pm$ 0.2 & 57.6 $\pm$ 0.5 & \texttt{RLLLLL}, \texttt{RLLRLL}, \texttt{RLRLLL} \\
\cmidrule(r){1-3} \cmidrule(l){4-7}
Chance rate & & & 0.0 & 0.0 & 0.0 & - \\
    \bottomrule
    \end{tabular}
        \caption{Word-order preferences of LMs. ``Lim.'' and ``Stx.'' indicate the working memory limitation and syntactic biases in the respective model architecture, respectively.}
        \label{tbl:results}
    \end{table*}

\begin{table*}[t]
    \centering
    \small
\fontsize{9}{8.5}\selectfont
\setlength{\tabcolsep}{2pt}
\begin{tabular}{llllrrrc}
\toprule
Model & Syntax & Lim. & Beam &  \multicolumn{1}{c}{Global $r$}$\uparrow$ & \multicolumn{1}{c}{Local $r$}$\uparrow$ & \texttt{L}-pref. $\rightarrow$ & Top3 langs. \\ 
\cmidrule(r){1-1} \cmidrule(r){2-2} \cmidrule(r){3-3}  \cmidrule(lr){4-4} \cmidrule(l){5-5} \cmidrule(l){6-6} \cmidrule(l){7-7} \cmidrule(l){8-8}
RNNG & TD & & & -22.6 $\pm$ 4.7 & 6.0 $\pm$ 14.5  & 14.5 $\pm$ 10.8 & \texttt{RLLRLL}, \texttt{RRRRLL}, \texttt{RRRLLL} \\
SRNNG & TD & \checkmark & & 2.0 $\pm$ 9.3 & 10.7 $\pm$ 7.8  & 10.2 $\pm$ 7.0 & \texttt{RLLRRR}, \texttt{RLRRRR}, \texttt{RLLRRL} \\
RNNG & LC & & & -17.6 $\pm$ 6.4 & 25.1 $\pm$ 13.4  & -21.2 $\pm$ 2.0 & \texttt{RRRRRL}, \texttt{RRLLRL}, \texttt{RRLRRL} \\
SRNNG & LC & \checkmark & & 19.0 $\pm$ 9.6 & 23.7 $\pm$ 13.0  & -40.6 $\pm$ 8.5 & \texttt{LRRRRR}, \texttt{LRLRRR}, \texttt{LLLRRR} \\
\cmidrule(r){1-4} \cmidrule(l){5-8}
RNNG & TD & & \checkmark & 9.4 $\pm$ 3.5 & -31.5 $\pm$ 11.6  & -30.1 $\pm$ 5.7 & \texttt{RRRRLL}, \texttt{RRRLLL}, \texttt{RRLLLL} \\
SRNNG & TD & \checkmark & \checkmark & 14.6 $\pm$ 8.7 & -2.8 $\pm$ 5.9  & -21.5 $\pm$ 8.9 & \texttt{LLLRRR}, \texttt{LLRRRR}, \texttt{RLRRRR} \\
RNNG & LC & & \checkmark & -23.4 $\pm$ 7.0 & 26.5 $\pm$ 13.9  & -26.2 $\pm$ 7.2 & \texttt{RLRLRL}, \texttt{RLRRLL}, \texttt{RRRRRL} \\
SRNNG & LC & \checkmark & \checkmark & 17.2 $\pm$ 8.5 & 18.3 $\pm$ 12.4  & -36.7 $\pm$ 9.7 & \texttt{LRRRRR}, \texttt{LRLRRR}, \texttt{RLLRRR} \\
\bottomrule
\end{tabular}

\caption{Comparson of the RNNG/SRNNG results with and without word-synchronous beam search}
\label{tbl:beam}
\end{table*}

\begin{table*}[t]
    \centering
\fontsize{9}{8}\selectfont
\setlength{\tabcolsep}{2pt}
    \begin{tabular}{lccrrrc}
    \toprule
    Model & Syntax & Lim. &  \multicolumn{1}{c}{Global $r$}$\uparrow$ & \multicolumn{1}{c}{Local $r$}$\uparrow$ & \texttt{L}-pref. $\rightarrow$ & Top3 langs. \\
\cmidrule(r){1-1} \cmidrule(r){2-2} \cmidrule(r){3-3}  \cmidrule(lr){4-4} \cmidrule(l){5-5} \cmidrule(l){6-6} \cmidrule(l){7-7}
Transformer & & & 12.4 $\pm$ 4.3 & 16.7 $\pm$ 6.4 & 24.1 $\pm$ 6.1 & \texttt{LLLLLL}, \texttt{LLRLLL}, \texttt{RLRLLL} \\
LSTM & \textcolor{gray}{\checkmark} & & 11.2 $\pm$ 14.7 & 27.0 $\pm$ 7.4 & -0.8 $\pm$ 11.3 & \texttt{RLRLLL}, \texttt{RLLLLL}, \texttt{RRRRRR} \\
SRN  & \checkmark & & 16.5 $\pm$ 9.4 & 38.4 $\pm$ 3.0 & -3.5 $\pm$ 10.4 & \texttt{RLLLLL}, \texttt{RLRLLL}, \texttt{RLRRLL} \\
\cmidrule(r){1-3} \cmidrule(l){4-7}
Word 5-gram & \checkmark & & 5.5 $\pm$ 1.0 & 17.1 $\pm$ 1.7 & -6.0 $\pm$ 1.6 & \texttt{RRLRRR}, \texttt{RRRRRR}, \texttt{LRLRRR} \\
Word 4-gram & \checkmark & & 6.6 $\pm$ 1.0 & 16.7 $\pm$ 1.6 & -6.7 $\pm$ 1.6 & \texttt{RRLRRR}, \texttt{RRRRRR}, \texttt{LRLRRR} \\
Word 3-gram & \checkmark & & 8.9 $\pm$ 0.7 & 17.7 $\pm$ 1.0 & -6.3 $\pm$ 1.0 & \texttt{RRRRRR}, \texttt{RRLRRR}, \texttt{LRLRRR} \\
\cmidrule(r){1-3} \cmidrule(l){4-7}
Trans. PLM & & TD & 30.5 $\pm$ 5.7 & 29.4 $\pm$ 9.1 & 36.2 $\pm$ 4.7 & \texttt{LLRRLL}, \texttt{LLRRRL}, \texttt{LLRLLL} \\
Trans. PLM & & LC & 30.2 $\pm$ 2.1 & 42.3 $\pm$ 1.1 & 35.7 $\pm$ 2.3 & \texttt{LLLLLL}, \texttt{LLRLLL}, \texttt{LLRRLL} \\
LSTM PLM & \textcolor{gray}{\checkmark} & TD & 11.9 $\pm$ 12.2 & 37.0 $\pm$ 7.3 & -27.2 $\pm$ 10.5 & \texttt{LRRRRL}, \texttt{LRLRRL}, \texttt{LRRRRR} \\
LSTM PLM & \textcolor{gray}{\checkmark} & LC & 23.6 $\pm$ 3.6 & 40.4 $\pm$ 2.5 & 0.6 $\pm$ 5.5 & \texttt{RLLRRR}, \texttt{LLLLLL}, \texttt{LLLRLL} \\
SRN PLM & \checkmark & TD & -5.4 $\pm$ 8.3 & 8.8 $\pm$ 7.8 & -53.8 $\pm$ 7.4 & \texttt{RLRRRR}, \texttt{LRLRRR}, \texttt{RLLRRR} \\
SRN PLM & \checkmark & LC & 9.3 $\pm$ 4.9 & 27.6 $\pm$ 10.5 & 2.8 $\pm$ 5.2 & \texttt{RLRRLL}, \texttt{RLRRRR}, \texttt{RLLRLL} \\
\cmidrule(r){1-3} \cmidrule(l){4-7}
5-gram PLM & \checkmark & TD & 11.8 $\pm$ 2.4 & 50.5 $\pm$ 2.8 & 10.2 $\pm$ 7.8 & \texttt{RLLRRL}, \texttt{RLRRRL}, \texttt{RLLLLL} \\
5-gram PLM & \checkmark & LC & 18.5 $\pm$ 0.9 & 47.0 $\pm$ 1.3 & 29.0 $\pm$ 1.4 & \texttt{LLLRLL}, \texttt{RLLRLL}, \texttt{RLRRLL} \\
4-gram PLM & \checkmark & TD & 29.2 $\pm$ 0.6 & 40.0 $\pm$ 1.6 & 4.3 $\pm$ 5.4 & \texttt{LLRRRL}, \texttt{RLRRRL}, \texttt{LLRRRR} \\
4-gram PLM & \checkmark & LC & 21.6 $\pm$ 0.6 & 50.5 $\pm$ 0.6 & 22.4 $\pm$ 0.9 & \texttt{RLLRLL}, \texttt{RLRRLL}, \texttt{LLLRLL} \\
3-gram PLM & \checkmark & TD & 19.9 $\pm$ 0.7 & 29.0 $\pm$ 1.8 & 17.3 $\pm$ 2.0 & \texttt{RLLRRL}, \texttt{LLLRRL}, \texttt{RLLRRR} \\
3-gram PLM & \checkmark & LC & 17.9 $\pm$ 0.2 & 55.7 $\pm$ 0.3 & 27.0 $\pm$ 0.5 & \texttt{RLLRRR}, \texttt{RLLRRL}, \texttt{RLRRRR} \\
\cmidrule(r){1-3} \cmidrule(l){4-7}
RNNG & & TD & -22.6 $\pm$ 4.7 & 6.0 $\pm$ 14.5 & 14.5 $\pm$ 10.8 & \texttt{RLLRLL}, \texttt{RRRRLL}, \texttt{RRRLLL} \\
RNNG & & LC & -17.6 $\pm$ 6.4 & 25.1 $\pm$ 13.4 & -21.1 $\pm$ 2.0 & \texttt{RRRRRL}, \texttt{RRLLRL}, \texttt{RRLRRL} \\
SRNNG & \checkmark & TD & 1.9 $\pm$ 9.3 & 10.7 $\pm$ 7.8 & 10.1 $\pm$ 7.0 & \texttt{RLLRRR}, \texttt{RLRRRR}, \texttt{RLLRRL} \\
SRNNG & \checkmark & LC & 19.1 $\pm$ 9.6 & 23.7 $\pm$ 13.0 & -40.7 $\pm$ 8.5 & \texttt{LRRRRR}, \texttt{LRLRRR}, \texttt{LLLRRR} \\
\cmidrule(r){1-3} \cmidrule(l){4-7}
LLaMA2 (7B) & &  & 6.9 $\pm$ 31.0 & 15.4 $\pm$ 2.5 & -4.6 $\pm$ 31.0 & \texttt{LRLLLL}, \texttt{LRRLLL}, \texttt{LRLRLL} \\
\bottomrule
\end{tabular}
\caption{The results of $\mathrm{PPL}^{1/2}$}
\label{tbl:results_k0.5}
\end{table*}

\begin{table*}[t]
    \centering
\fontsize{9}{8}\selectfont
\setlength{\tabcolsep}{2pt}
    \begin{tabular}{lccrrrc}
    \toprule
    Model & Syntax & Lim. &  \multicolumn{1}{c}{Global $r$}$\uparrow$ & \multicolumn{1}{c}{Local $r$}$\uparrow$ & \texttt{L}-pref. $\rightarrow$ & Top3 langs. \\
\cmidrule(r){1-1} \cmidrule(r){2-2} \cmidrule(r){3-3}  \cmidrule(lr){4-4} \cmidrule(l){5-5} \cmidrule(l){6-6} \cmidrule(l){7-7}
Transformer & & & 11.6 $\pm$ 4.3 & 16.5 $\pm$ 6.4 & 23.1 $\pm$ 6.1 & \texttt{LLLLLL}, \texttt{LLRLLL}, \texttt{RLRLLL} \\
LSTM & \textcolor{gray}{\checkmark} & & 9.8 $\pm$ 14.7 & 26.6 $\pm$ 7.4 & -1.8 $\pm$ 11.3 & \texttt{RLRLLL}, \texttt{RLLLLL}, \texttt{RRRRRR} \\
SRN  & \checkmark & & 16.1 $\pm$ 9.4 & 38.3 $\pm$ 3.0 & -3.6 $\pm$ 10.4 & \texttt{RLLLLL}, \texttt{RLRLLL}, \texttt{RLRRLL} \\
\cmidrule(r){1-3} \cmidrule(l){4-7}
Word 5-gram & \checkmark & & 5.4 $\pm$ 1.0 & 16.7 $\pm$ 1.7 & -5.3 $\pm$ 1.6 & \texttt{RRLRRR}, \texttt{RRRRRR}, \texttt{LRLRRR} \\
Word 4-gram & \checkmark & & 6.4 $\pm$ 1.0 & 16.1 $\pm$ 1.6 & -5.8 $\pm$ 1.6 & \texttt{RRLRRR}, \texttt{RRRRRR}, \texttt{LRLRRR} \\
Word 3-gram & \checkmark & & 8.5 $\pm$ 0.7 & 17.1 $\pm$ 1.0 & -5.6 $\pm$ 1.0 & \texttt{RRRRRR}, \texttt{RRLRRR}, \texttt{LRLRRR} \\
\cmidrule(r){1-3} \cmidrule(l){4-7}
Trans. PLM & & TD & 30.3 $\pm$ 5.7 & 29.7 $\pm$ 9.1 & 35.3 $\pm$ 4.7 & \texttt{LLRRLL}, \texttt{LLRRRL}, \texttt{LLRLLL} \\
Trans. PLM & & LC & 30.3 $\pm$ 2.1 & 42.5 $\pm$ 1.1 & 34.1 $\pm$ 2.3 & \texttt{LLLLLL}, \texttt{LLRLLL}, \texttt{LLRRLL} \\
\cmidrule(r){1-3} \cmidrule(l){4-7}
LSTM PLM & \textcolor{gray}{\checkmark} & TD & 11.8 $\pm$ 12.2 & 37.0 $\pm$ 7.3 & -27.1 $\pm$ 10.5 & \texttt{LRRRRL}, \texttt{LRLRRL}, \texttt{LRRRRR} \\
LSTM PLM & \textcolor{gray}{\checkmark} & LC & 23.6 $\pm$ 3.6 & 40.4 $\pm$ 2.5 & 0.4 $\pm$ 5.5 & \texttt{RLLRRR}, \texttt{LLLLLL}, \texttt{LLLRLL} \\
SRN PLM & \checkmark & TD & -5.5 $\pm$ 8.3 & 8.7 $\pm$ 7.8 & -53.7 $\pm$ 7.4 & \texttt{RLRRRR}, \texttt{LRLRRR}, \texttt{RLLRRR} \\
SRN PLM & \checkmark & LC & 9.9 $\pm$ 4.9 & 27.4 $\pm$ 10.5 & 2.8 $\pm$ 5.2 & \texttt{RLRRLL}, \texttt{RLRRRR}, \texttt{RLLRLL} \\
\cmidrule(r){1-3} \cmidrule(l){4-7}
5-gram PLM & \checkmark & TD & 11.9 $\pm$ 2.4 & 50.4 $\pm$ 2.8 & 10.2 $\pm$ 7.8 & \texttt{RLLRRL}, \texttt{RLRRRL}, \texttt{RLLLLL} \\
5-gram PLM & \checkmark & LC & 18.7 $\pm$ 0.9 & 47.1 $\pm$ 1.3 & 28.5 $\pm$ 1.4 & \texttt{LLLRLL}, \texttt{RLLRLL}, \texttt{RLRRLL} \\
4-gram PLM & \checkmark & TD & 29.2 $\pm$ 0.6 & 40.0 $\pm$ 1.6 & 4.4 $\pm$ 5.4 & \texttt{LLRRRL}, \texttt{RLRRRL}, \texttt{LLRRRR} \\
4-gram PLM & \checkmark & LC & 21.9 $\pm$ 0.6 & 50.6 $\pm$ 0.6 & 22.0 $\pm$ 0.9 & \texttt{RLLRLL}, \texttt{RLRRLL}, \texttt{LLLRLL} \\
3-gram PLM & \checkmark & TD & 19.9 $\pm$ 0.7 & 29.0 $\pm$ 1.8 & 17.3 $\pm$ 2.0 & \texttt{RLLRRL}, \texttt{LLLRRL}, \texttt{RLLRRR} \\
3-gram PLM & \checkmark & LC & 18.2 $\pm$ 0.2 & 55.6 $\pm$ 0.3 & 26.9 $\pm$ 0.5 & \texttt{RLLRRR}, \texttt{RLLRRL}, \texttt{RLRRRR} \\
\cmidrule(r){1-3} \cmidrule(l){4-7}
RNNG & & TD & -22.6 $\pm$ 4.7 & 6.0 $\pm$ 14.5 & 14.5 $\pm$ 10.8 & \texttt{RLLRLL}, \texttt{RRRRLL}, \texttt{RRRLLL} \\
RNNG & & LC & -17.6 $\pm$ 6.4 & 25.1 $\pm$ 13.4 & -21.2 $\pm$ 2.0 & \texttt{RRRRRL}, \texttt{RRLLRL}, \texttt{RRLRRL} \\
SRNNG & \checkmark & TD & 2.1 $\pm$ 9.3 & 10.7 $\pm$ 7.8 & 10.2 $\pm$ 7.0 & \texttt{RLLRRR}, \texttt{RLRRRR}, \texttt{RLLRRL} \\
SRNNG & \checkmark & LC & 19.0 $\pm$ 9.6 & 23.6 $\pm$ 13.0 & -40.4 $\pm$ 8.5 & \texttt{LRRRRR}, \texttt{LRLRRR}, \texttt{LLLRRR} \\
\cmidrule(r){1-3} \cmidrule(l){4-7}
LLaMA2 (7B) & &  & 6.9 $\pm$ 31.0 & 15.5 $\pm$ 2.5 & -4.8 $\pm$ 31.0 & \texttt{LRLLLL}, \texttt{LRRLLL}, \texttt{LRLRLL} \\
\bottomrule
\end{tabular}
\caption{The results of $\mathrm{PPL}^{2}$}
\label{tbl:results_k2}
\end{table*}

\begin{table*}[t]
    \centering
\fontsize{9}{8}\selectfont
\setlength{\tabcolsep}{2pt}
    \begin{tabular}{lccrrrc}
    \toprule
    Model & Syntax & Lim. &  \multicolumn{1}{c}{Global $r$}$\uparrow$ & \multicolumn{1}{c}{Local $r$}$\uparrow$ & \texttt{L}-pref. $\rightarrow$ & Top3 langs. \\
\cmidrule(r){1-1} \cmidrule(r){2-2} \cmidrule(r){3-3}  \cmidrule(lr){4-4} \cmidrule(l){5-5} \cmidrule(l){6-6} \cmidrule(l){7-7}
Transformer & & & 11.1 $\pm$ 4.3 & 16.3 $\pm$ 6.4 & 22.5 $\pm$ 6.1 & \texttt{LLLLLL}, \texttt{LLRLLL}, \texttt{RLRLLL} \\
LSTM & \textcolor{gray}{\checkmark} & & 9.2 $\pm$ 14.7 & 26.4 $\pm$ 7.4 & -2.3 $\pm$ 11.3 & \texttt{RLRLLL}, \texttt{RLLLLL}, \texttt{RRRRRR} \\
SRN  & \checkmark & & 16.1 $\pm$ 9.4 & 38.3 $\pm$ 3.0 & -3.5 $\pm$ 10.4 & \texttt{RLLLLL}, \texttt{RLRLLL}, \texttt{RLRRLL} \\
Word 5-gram & \checkmark & & 5.3 $\pm$ 1.0 & 16.3 $\pm$ 1.7 & -4.8 $\pm$ 1.6 & \texttt{RRLRRR}, \texttt{RRRRRR}, \texttt{LRLRRR} \\
Word 4-gram & \checkmark & & 6.3 $\pm$ 1.0 & 15.7 $\pm$ 1.6 & -5.3 $\pm$ 1.6 & \texttt{RRLRRR}, \texttt{RRRRRR}, \texttt{LRLRRR} \\
Word 3-gram & \checkmark & & 8.3 $\pm$ 0.7 & 16.8 $\pm$ 1.0 & -5.1 $\pm$ 1.0 & \texttt{RRRRRR}, \texttt{RRLRRR}, \texttt{LRLRRR} \\
\cmidrule(r){1-3} \cmidrule(l){4-7}
Trans. PLM & & TD & 30.2 $\pm$ 5.7 & 29.9 $\pm$ 9.1 & 34.7 $\pm$ 4.7 & \texttt{LLRRLL}, \texttt{LLRRRL}, \texttt{LLRLLL} \\
Trans. PLM & & LC & 30.3 $\pm$ 2.1 & 42.6 $\pm$ 1.1 & 33.0 $\pm$ 2.3 & \texttt{LLLLLL}, \texttt{LLRLLL}, \texttt{LLRRLL} \\
LSTM PLM & \textcolor{gray}{\checkmark} & TD & 11.8 $\pm$ 12.2 & 36.9 $\pm$ 7.3 & -27.0 $\pm$ 10.5 & \texttt{LRRRRL}, \texttt{LRLRRL}, \texttt{LRRRRR} \\
LSTM PLM & \textcolor{gray}{\checkmark} & LC & 23.6 $\pm$ 3.6 & 40.3 $\pm$ 2.5 & 0.2 $\pm$ 5.5 & \texttt{RLLRRR}, \texttt{LLLLLL}, \texttt{LLLRLL} \\
SRN PLM & \checkmark & TD & -5.7 $\pm$ 8.3 & 8.6 $\pm$ 7.8 & -53.6 $\pm$ 7.4 & \texttt{RLRRRR}, \texttt{LRLRRR}, \texttt{RLLRRR} \\
SRN PLM & \checkmark & LC & 10.2 $\pm$ 4.9 & 27.3 $\pm$ 10.5 & 2.7 $\pm$ 5.2 & \texttt{RLRRLL}, \texttt{RLRRRR}, \texttt{RLLRLL} \\
\cmidrule(r){1-3} \cmidrule(l){4-7}
5-gram PLM & \checkmark & TD & 11.9 $\pm$ 2.4 & 50.3 $\pm$ 2.8 & 10.1 $\pm$ 7.8 & \texttt{RLLRRL}, \texttt{RLRRRL}, \texttt{RLLLLL} \\
5-gram PLM & \checkmark & LC & 18.8 $\pm$ 0.9 & 47.2 $\pm$ 1.3 & 28.2 $\pm$ 1.4 & \texttt{LLLRLL}, \texttt{RLLRLL}, \texttt{RLRRLL} \\
4-gram PLM & \checkmark & TD & 29.2 $\pm$ 0.6 & 40.0 $\pm$ 1.6 & 4.5 $\pm$ 5.4 & \texttt{LLRRRL}, \texttt{RLRRRL}, \texttt{LLRRRR} \\
4-gram PLM & \checkmark & LC & 22.1 $\pm$ 0.6 & 50.6 $\pm$ 0.6 & 21.7 $\pm$ 0.9 & \texttt{RLLRLL}, \texttt{RLRRLL}, \texttt{LLLRLL} \\
3-gram PLM & \checkmark & TD & 19.9 $\pm$ 0.7 & 29.0 $\pm$ 1.8 & 17.4 $\pm$ 2.0 & \texttt{RLLRRL}, \texttt{LLLRRL}, \texttt{RLLRRR} \\
3-gram PLM & \checkmark & LC & 18.4 $\pm$ 0.2 & 55.6 $\pm$ 0.3 & 26.8 $\pm$ 0.5 & \texttt{RLLRRR}, \texttt{RLLRRL}, \texttt{RLRRRR} \\
\cmidrule(r){1-3} \cmidrule(l){4-7}
RNNG & & TD & -22.5 $\pm$ 4.7 & 6.0 $\pm$ 14.5 & 14.5 $\pm$ 10.8 & \texttt{RLLRLL}, \texttt{RRRRLL}, \texttt{RRRLLL} \\
RNNG & & LC & -17.6 $\pm$ 6.4 & 25.1 $\pm$ 13.4 & -21.2 $\pm$ 2.0 & \texttt{RRRRRL}, \texttt{RRLLRL}, \texttt{RRLRRL} \\
SRNNG & \checkmark & TD & 2.2 $\pm$ 9.3 & 10.7 $\pm$ 7.8 & 10.2 $\pm$ 7.0 & \texttt{RLLRRR}, \texttt{RLRRRR}, \texttt{RLLRRL} \\
SRNNG & \checkmark & LC & 18.9 $\pm$ 9.6 & 23.6 $\pm$ 13.0 & -40.2 $\pm$ 8.5 & \texttt{LRRRRR}, \texttt{LRLRRR}, \texttt{LLLRRR} \\
\cmidrule(r){1-3} \cmidrule(l){4-7}
LLaMA2 (7B) & &  & 6.9 $\pm$ 31.0 & 15.5 $\pm$ 2.5 & -4.9 $\pm$ 31.0 & \texttt{LRLLLL}, \texttt{LRRLLL}, \texttt{LRLRLL} \\
\bottomrule
\end{tabular}
\caption{The results of $\mathrm{PPL}^{3}$}
\label{tbl:results_k3}
\end{table*}

\begin{table*}[t]
    \centering
\fontsize{9}{8}\selectfont
\setlength{\tabcolsep}{2pt}
    \begin{tabular}{lccrrrc}
    \toprule
    Model & Syntax & Lim. &  \multicolumn{1}{c}{Global $r$}$\uparrow$ & \multicolumn{1}{c}{Local $r$}$\uparrow$ & \texttt{L}-pref. $\rightarrow$ & Top3 langs. \\
\cmidrule(r){1-1} \cmidrule(r){2-2} \cmidrule(r){3-3}  \cmidrule(lr){4-4} \cmidrule(l){5-5} \cmidrule(l){6-6} \cmidrule(l){7-7}
Transformer & & & 12.6 $\pm$ 4.3 & 16.8 $\pm$ 6.4 & 24.4 $\pm$ 6.1 & \texttt{LLLLLL}, \texttt{LLRLLL}, \texttt{RLRLLL} \\
LSTM & \textcolor{gray}{\checkmark} & & 11.8 $\pm$ 14.7 & 27.2 $\pm$ 7.4 & -0.4 $\pm$ 11.3 & \texttt{RLRLLL}, \texttt{RLLLLL}, \texttt{RRRRRR} \\
SRN  & \checkmark & & 16.7 $\pm$ 9.4 & 38.4 $\pm$ 3.0 & -3.5 $\pm$ 10.4 & \texttt{RLLLLL}, \texttt{RLRLLL}, \texttt{RLRRLL} \\
\cmidrule(r){1-3} \cmidrule(l){4-7}
Word 5-gram & \checkmark & & 5.5 $\pm$ 1.0 & 17.3 $\pm$ 1.7 & -6.3 $\pm$ 1.6 & \texttt{RRLRRR}, \texttt{RRRRRR}, \texttt{LRLRRR} \\
Word 4-gram & \checkmark & & 6.7 $\pm$ 1.0 & 16.8 $\pm$ 1.6 & -7.0 $\pm$ 1.6 & \texttt{RRLRRR}, \texttt{RRRRRR}, \texttt{LRLRRR} \\
Word 3-gram & \checkmark & & 9.0 $\pm$ 0.7 & 17.9 $\pm$ 1.0 & -6.6 $\pm$ 1.0 & \texttt{RRRRRR}, \texttt{RRLRRR}, \texttt{LRLRRR} \\
\cmidrule(r){1-3} \cmidrule(l){4-7}
Trans. PLM & & TD & 30.5 $\pm$ 5.7 & 29.4 $\pm$ 9.1 & 36.5 $\pm$ 4.7 & \texttt{LLRRLL}, \texttt{LLRRRL}, \texttt{LLRLLL} \\
Trans. PLM & & LC & 30.2 $\pm$ 2.1 & 42.2 $\pm$ 1.1 & 36.2 $\pm$ 2.3 & \texttt{LLLLLL}, \texttt{LLRLLL}, \texttt{LLRRLL} \\
LSTM PLM & \textcolor{gray}{\checkmark} & TD & 12.0 $\pm$ 12.2 & 37.1 $\pm$ 7.3 & -27.2 $\pm$ 10.5 & \texttt{LRRRRL}, \texttt{LRLRRL}, \texttt{LRRRRR} \\
LSTM PLM & \textcolor{gray}{\checkmark} & LC & 23.6 $\pm$ 3.6 & 40.4 $\pm$ 2.5 & 0.7 $\pm$ 5.5 & \texttt{RLLRRR}, \texttt{LLLLLL}, \texttt{LLLRLL} \\
SRN PLM & \checkmark & TD & -5.3 $\pm$ 8.3 & 8.8 $\pm$ 7.8 & -53.8 $\pm$ 7.4 & \texttt{RLRRRR}, \texttt{LRLRRR}, \texttt{RLLRRR} \\
SRN PLM & \checkmark & LC & 9.1 $\pm$ 4.9 & 27.6 $\pm$ 10.5 & 2.8 $\pm$ 5.2 & \texttt{RLRRLL}, \texttt{RLRRRR}, \texttt{RLLRLL} \\
\cmidrule(r){1-3} \cmidrule(l){4-7}
5-gram PLM & \checkmark & TD & 11.8 $\pm$ 2.4 & 50.5 $\pm$ 2.8 & 10.2 $\pm$ 7.8 & \texttt{RLLRRL}, \texttt{RLRRRL}, \texttt{RLLLLL} \\
5-gram PLM & \checkmark & LC & 18.4 $\pm$ 0.9 & 46.9 $\pm$ 1.3 & 29.1 $\pm$ 1.4 & \texttt{LLLRLL}, \texttt{RLLRLL}, \texttt{RLRRLL} \\
4-gram PLM & \checkmark & TD & 29.2 $\pm$ 0.6 & 40.0 $\pm$ 1.6 & 4.3 $\pm$ 5.4 & \texttt{LLRRRL}, \texttt{RLRRRL}, \texttt{LLRRRR} \\
4-gram PLM & \checkmark & LC & 21.5 $\pm$ 0.6 & 50.5 $\pm$ 0.6 & 22.5 $\pm$ 0.9 & \texttt{RLLRLL}, \texttt{RLRRLL}, \texttt{LLLRLL} \\
3-gram PLM & \checkmark & TD & 19.9 $\pm$ 0.7 & 29.0 $\pm$ 1.8 & 17.2 $\pm$ 2.0 & \texttt{RLLRRL}, \texttt{LLLRRL}, \texttt{RLLRRR} \\
3-gram PLM & \checkmark & LC & 17.8 $\pm$ 0.2 & 55.7 $\pm$ 0.3 & 27.0 $\pm$ 0.5 & \texttt{RLLRRR}, \texttt{RLLRRL}, \texttt{RLRRRR} \\
\cmidrule(r){1-3} \cmidrule(l){4-7}
RNNG & & TD & -22.7 $\pm$ 4.7 & 6.0 $\pm$ 14.5 & 14.5 $\pm$ 10.8 & \texttt{RLLRLL}, \texttt{RRRRLL}, \texttt{RRRLLL} \\
RNNG & & LC & -17.6 $\pm$ 6.4 & 25.1 $\pm$ 13.4 & -21.1 $\pm$ 2.0 & \texttt{RRRRRL}, \texttt{RRLLRL}, \texttt{RRLRRL} \\
SRNNG & \checkmark & TD & 1.8 $\pm$ 9.3 & 10.7 $\pm$ 7.8 & 10.1 $\pm$ 7.0 & \texttt{RLLRRR}, \texttt{RLRRRR}, \texttt{RLLRRL} \\
SRNNG & \checkmark & LC & 19.1 $\pm$ 9.6 & 23.8 $\pm$ 13.0 & -40.7 $\pm$ 8.5 & \texttt{LRRRRR}, \texttt{LRLRRR}, \texttt{LLLRRR} \\
\cmidrule(r){1-3} \cmidrule(l){4-7}
LLaMA2 (7B) & &  & 6.8 $\pm$ 31.0 & 15.4 $\pm$ 2.5 & -4.5 $\pm$ 31.0 & \texttt{LRLLLL}, \texttt{LRRLLL}, \texttt{LRLRLL} \\
\bottomrule
\end{tabular}
\caption{The results of $\log \mathrm{PPL}$}
\label{tbl:results_klog}
\end{table*}

\begin{table*}[t]
    \centering
    \small
\fontsize{7}{7}\selectfont
\setlength{\tabcolsep}{3pt}
\begin{tabular}{rll}
\toprule
Probability & Production rule & Relevant parameter \joystick \\
\cmidrule(r){1-1} \cmidrule(r){2-2} \cmidrule(l){3-3}
1 & ROOT $\rightarrow$ S & \\ 
1/2 & S $\rightarrow$ NP\_Subj\_S VP\_S	 & $\joystick^{\boldsymbol \Ss}$ \\ 
1/2 & S $\rightarrow$ NP\_Subj\_P VP\_P	 & $\joystick^{\boldsymbol \Ss}$ \\ 
1/3 & VP\_S $\rightarrow$ VP\_Past\_S & \\ 
1/3 & VP\_S $\rightarrow$ VP\_Pres\_S & \\ 
1/3 & VP\_S $\rightarrow$ VP\_Comp\_S & \\ 
1/3 & VP\_P $\rightarrow$ VP\_Past\_P & \\ 
1/3 & VP\_P $\rightarrow$ VP\_Pres\_P & \\ 
1/3 & VP\_P $\rightarrow$ VP\_Comp\_P & \\ 
1/2 & VP\_Comp\_S $\rightarrow$ VP\_Comp\_Pres\_S & \\ 
1/2 & VP\_Comp\_S $\rightarrow$ VP\_Comp\_Past\_S & \\ 
1/2 & VP\_Comp\_P $\rightarrow$ VP\_Comp\_Pres\_P & \\ 
1/2 & VP\_Comp\_P $\rightarrow$ VP\_Comp\_Past\_P & \\ 
1/2 & VP\_Past\_S $\rightarrow$ IVerb\_Past\_S & \\ 
1/2 & VP\_Past\_S $\rightarrow$ NP\_Obj TVerb\_Past\_S	 & $\joystick^{\boldsymbol \VP}$ \\ 
1/2 & VP\_Pres\_S $\rightarrow$ IVerb\_Pres\_S & \\ 
1/2 & VP\_Pres\_S $\rightarrow$ NP\_Obj TVerb\_Pres\_S	 & $\joystick^{\boldsymbol \VP}$ \\ 
1/2 & VP\_Past\_P $\rightarrow$ IVerb\_Past\_P & \\ 
1/2 & VP\_Past\_P $\rightarrow$ NP\_Obj TVerb\_Past\_P	 & $\joystick^{\boldsymbol \VP}$ \\ 
1/2 & VP\_Pres\_P $\rightarrow$ IVerb\_Pres\_P & \\ 
1/2 & VP\_Pres\_P $\rightarrow$ NP\_Obj TVerb\_Pres\_P	 & $\joystick^{\boldsymbol \VP}$ \\ 
1 & VP\_Comp\_Pres\_S $\rightarrow$ S\_Comp Verb\_Comp\_Pres\_S	 & $\joystick^{\boldsymbol \VP}$ \\ 
1 & VP\_Comp\_Past\_S $\rightarrow$ S\_Comp Verb\_Comp\_Past\_S	 & $\joystick^{\boldsymbol \VP}$ \\ 
1 & VP\_Comp\_Pres\_P $\rightarrow$ S\_Comp Verb\_Comp\_Pres\_P	 & $\joystick^{\boldsymbol \VP}$ \\ 
1 & VP\_Comp\_Past\_P $\rightarrow$ S\_Comp Verb\_Comp\_Past\_P	 & $\joystick^{\boldsymbol \VP}$ \\ 
1 & S\_Comp $\rightarrow$ S Comp	 & $\joystick^{\boldsymbol \Comp}$ \\ 
1 & NP\_Subj\_S $\rightarrow$ NP\_S Subj	 & $\joystick^{\boldsymbol \Case}$ \\ 
1 & NP\_Subj\_P $\rightarrow$ NP\_P Subj	 & $\joystick^{\boldsymbol \Case}$ \\ 
1/2 & NP\_Obj $\rightarrow$ NP\_S Obj	 & $\joystick^{\boldsymbol \Case}$ \\ 
1/2 & NP\_Obj $\rightarrow$ NP\_P Obj	 & $\joystick^{\boldsymbol \Case}$ \\ 
5/21 & NP\_S $\rightarrow$ Noun\_S & \\ 
5/21 & NP\_S $\rightarrow$ Adj Noun\_S	 & $\joystick^{\boldsymbol \NP}$ \\ 
5/21 & NP\_S $\rightarrow$ VP\_S Rel Noun\_S	 & $\joystick^{\boldsymbol \Rel}$ \\ 
5/21 & NP\_S $\rightarrow$ Pronoun\_S & \\ 
1/21 & NP\_S $\rightarrow$ PP NP\_S	 & $\joystick^{\boldsymbol \PP}$ \\ 
10/43 & NP\_P $\rightarrow$ Noun\_P & \\ 
10/43 & NP\_P $\rightarrow$ Adj Noun\_P	 & $\joystick^{\boldsymbol \NP}$ \\ 
10/43 & NP\_P $\rightarrow$ VP\_P Rel Noun\_P	 & $\joystick^{\boldsymbol \Rel}$ \\ 
10/43 & NP\_P $\rightarrow$ Pronoun\_P & \\ 
2/43 & NP\_P $\rightarrow$ PP NP\_P	 & $\joystick^{\boldsymbol \PP}$ \\ 
1/172 & NP\_P $\rightarrow$ NP\_S CC NP\_S & \\ 
1/172 & NP\_P $\rightarrow$ NP\_P CC NP\_P & \\ 
1/172 & NP\_P $\rightarrow$ NP\_P CC NP\_S & \\ 
1/172 & NP\_P $\rightarrow$ NP\_S CC NP\_P & \\ 
1/2 & PP $\rightarrow$ NP\_S Prep	 & $\joystick^{\boldsymbol \PP}$ \\ 
1/2 & PP $\rightarrow$ NP\_P Prep	 & $\joystick^{\boldsymbol \PP}$ \\ 
1/43 & Adj $\rightarrow$ Adj CC Adj & \\ 
1/566 & TVerb\_Past\_S $\rightarrow$ TVerb\_Past\_S CC TVerb\_Past\_S & \\ 
1/566 & TVerb\_Pres\_S $\rightarrow$ TVerb\_Pres\_S CC TVerb\_Pres\_S & \\ 
1/566 & IVerb\_Past\_S $\rightarrow$ IVerb\_Past\_S CC IVerb\_Past\_S & \\ 
1/566 & IVerb\_Pres\_S $\rightarrow$ IVerb\_Pres\_S CC IVerb\_Pres\_S & \\ 
1/566 & TVerb\_Past\_P $\rightarrow$ TVerb\_Past\_P CC TVerb\_Past\_P & \\ 
1/566 & TVerb\_Pres\_P $\rightarrow$ TVerb\_Pres\_P CC TVerb\_Pres\_P & \\ 
1/566 & IVerb\_Past\_P $\rightarrow$ IVerb\_Past\_P CC IVerb\_Past\_P & \\ 
1/566 & IVerb\_Pres\_P $\rightarrow$ IVerb\_Pres\_P CC IVerb\_Pres\_P & \\ 
\cmidrule(r){1-1} \cmidrule(r){2-2} \cmidrule(r){3-3}
1 & Verb\_Comp\_Past\_S $\rightarrow$ word $\sim$ Dict[Verb\_Comp\_Past\_S] \ \ \# 22 types  & \\
1 & Verb\_Comp\_Past\_P $\rightarrow$ word $\sim$ Dict[Verb\_Comp\_Past\_P] \ \ \# 22 types  & \\
565/566 & IVerb\_Past\_S $\rightarrow$ word $\sim$ Dict[IVerb\_Past\_S] \ \ \# 113 types  & \\
565/566 & IVerb\_Past\_P $\rightarrow$ word $\sim$ Dict[IVerb\_Past\_P] \ \ \# 113 types  & \\
565/566 & TVerb\_Past\_S $\rightarrow$ word $\sim$ Dict[TVerb\_Past\_S] \ \ \# 113 types  & \\
565/566 & TVerb\_Past\_P $\rightarrow$ word $\sim$ Dict[TVerb\_Past\_P] \ \ \# 113 types  & \\
1 & Verb\_Comp\_Pres\_S $\rightarrow$ word $\sim$ Dict[Verb\_Comp\_Pres\_S] \ \ \# 22 types  & \\
1 & Verb\_Comp\_Pres\_P $\rightarrow$ word $\sim$ Dict[Verb\_Comp\_Pres\_P] \ \ \# 22 types  & \\
565/566 & IVerb\_Pres\_S $\rightarrow$ word $\sim$ Dict[IVerb\_Pres\_S] \ \ \# 113 types  & \\
565/566 & IVerb\_Pres\_P $\rightarrow$ word $\sim$ Dict[IVerb\_Pres\_P] \ \ \# 113 types  & \\
565/566 & TVerb\_Pres\_S $\rightarrow$ word $\sim$ Dict[TVerb\_Pres\_S] \ \ \# 113 types  & \\
565/566 & TVerb\_Pres\_P $\rightarrow$ word $\sim$ Dict[TVerb\_Pres\_P] \ \ \# 113 types  & \\
1 & Noun\_S $\rightarrow$ word $\sim$ Dict[Noun\_S] \ \ \# 162 types  & \\
1 & Noun\_P $\rightarrow$ word $\sim$ Dict[Noun\_P] \ \ \# 162 types  & \\
1 & Pronoun\_S $\rightarrow$ word $\sim$ Dict[Pronoun\_S] \ \ \# 5 types  & \\
1 & Pronoun\_P $\rightarrow$ word $\sim$ Dict[Pronoun\_P] \ \ \# 2 types  & \\
42/43 & Adj $\rightarrow$ word $\sim$ Dict[Adj] \ \ \# 42 types  & \\
1 & Prep $\rightarrow$ word $\sim$ Dict[Prep] \ \ \# 4 types  & \\
1 & CC $\rightarrow$ da & \\
1 & Comp $\rightarrow$ sa & \\
1 & Rel $\rightarrow$ rel & \\
1 & Subj $\rightarrow$ sub & \\
1 & Obj $\rightarrow$ ob & \\
\bottomrule
\end{tabular}

\caption{The base grammar we used to create artificial language data. The relevant switch in the third column overwrites the linearization order in the corresponding rule. The lexical items are randomly sampled from the pseudoword dictionary.}
\label{tbl:grammar}
\end{table*}

\begin{table*}[ht]
    \centering
    \small
\begin{minipage}[t]{\hsize}
\renewcommand{\arraystretch}{0.4}
    \centering
    \begin{tabular}{p{3cm}p{5cm}p{4.5cm}} \toprule
     \multirow{10}{1cm}{Fairseq model}
      & share-decoder-input-output-embed & True \\
      & embed\_dim & 128 \\
      & ffn\_embed\_dim & 512 \\
      & layers & 2 \\
      & heads & 2 \\
      & dropout & 0.3 \\
      & attention\_dropout & 0.1 \\
      & \#params. & 462K \\
    \cmidrule(lr){1-1} \cmidrule(lr){2-2} \cmidrule(lr){3-3}
    \multirow{5}{*}{Optimizer} & algorithm & AdamW \\
    & learning rates & 5e-4 \\
    & betas & (0.9, 0.98) \\
    & weight decay & 0.01 \\
    & clip norm & 0.0 \\
    \cmidrule(lr){1-1} \cmidrule(lr){2-2} \cmidrule(lr){3-3}
    \multirow{3}{3cm}{Learning rate scheduler} & type & inverse\_sqrt \\
    & warmup updates & 400 \\
    & warmup init learning rate & 1e-7 \\
    \cmidrule(lr){1-1} \cmidrule(lr){2-2} \cmidrule(lr){3-3}
    \multirow{3}{*}{Training} & batch size & 512 tokens \\
    & sample-break-mode & none \\
    & epochs & 10 \\ 
    \bottomrule
        \end{tabular}
        \subcaption{Transformer.}
        \label{tbl:hyperparam_tl}
        \vspace{0.2cm}
\end{minipage}

\begin{minipage}[t]{\hsize}
\renewcommand{\arraystretch}{0.4}
    \centering
    \begin{tabular}{p{3cm}p{5cm}p{4.5cm}} \toprule
     \multirow{8}{1cm}{Fairseq model} 
      & share-decoder-input-output-embed & True \\
      & embed\_dim & 128 \\
      & hiden\_size & 512 \\
      & layers & 2 \\
      & dropout & 0.1 \\
      & \#params. & 3,547K \\
    \cmidrule(lr){1-1} \cmidrule(lr){2-2} \cmidrule(lr){3-3}
    \multirow{5}{*}{Optimizer} & algorithm & AdamW \\
    & learning rates & 5e-4 \\
    & betas & (0.9, 0.98) \\
    & weight decay & 0.01 \\
    & clip norm & 0.0 \\
    \cmidrule(lr){1-1} \cmidrule(lr){2-2} \cmidrule(lr){3-3}
    \multirow{3}{3cm}{Learning rate scheduler} & type & inverse\_sqrt \\
    & warmup updates & 400 \\
    & warmup init learning rate & 1e-7 \\
    \cmidrule(lr){1-1} \cmidrule(lr){2-2} \cmidrule(lr){3-3}
    \multirow{3}{3cm}{Training} & batch size & 512 tokens \\
    & sample-break-mode & none \\ 
    & epochs & 10 \\ \bottomrule
        \end{tabular}
        \subcaption{LSTM.}
        \label{tbl:hyperparam_lstm}
        \vspace{0.2cm}
\end{minipage}

\begin{minipage}[t]{\hsize}
\renewcommand{\arraystretch}{0.4}
    \centering
    \begin{tabular}{p{3cm}p{5cm}p{4.5cm}} \toprule
     \multirow{8}{1cm}{Fairseq model} 
      & share-decoder-input-output-embed & True \\
      & embed\_dim & 64 \\
      & hiden\_size & 64 \\
      & layers & 2 \\
      & dropout & 0.1 \\
      & \#params. & 49K \\
    \cmidrule(lr){1-1} \cmidrule(lr){2-2} \cmidrule(lr){3-3}
    \multirow{5}{*}{Optimizer} & algorithm & AdamW \\
    & learning rates & 5e-4 \\
    & betas & (0.9, 0.98) \\
    & weight decay & 0.01 \\
    & clip norm & 0.0 \\
    \cmidrule(lr){1-1} \cmidrule(lr){2-2} \cmidrule(lr){3-3}
    \multirow{3}{3cm}{Learning rate scheduler} & type & inverse\_sqrt \\
    & warmup updates & 400 \\
    & warmup init learning rate & 1e-7 \\
    \cmidrule(lr){1-1} \cmidrule(lr){2-2} \cmidrule(lr){3-3}
    \multirow{3}{3cm}{Training} & batch size & 512 tokens \\
    & sample-break-mode & none \\ 
    & epochs & 10 \\ \bottomrule
        \end{tabular}
        \subcaption{SRN.}
        \label{tbl:hyperparam_rnn}
        \vspace{0.2cm}
\end{minipage}

\caption{Hyperparameters of standard LMs and PLMs}
\label{tbl:hyper_params}
\end{table*}

\begin{table*}[ht]
    \centering
    \small
\begin{minipage}[t]{\hsize}
\renewcommand{\arraystretch}{0.5}
    \centering
    \begin{tabular}{p{3cm}p{5cm}p{4.5cm}} \toprule
     \multirow{7}{1cm}{model} & composition & BiLSTM \\
      & recurrence & LSTM \\
      & embed\_dim & 256 \\
      & hiden\_size & 256 \\
      & layers & 2 \\
      & dropout & 0.3 \\
      & \#params. & 2,440K \\
    \cmidrule(lr){1-1} \cmidrule(lr){2-2} \cmidrule(lr){3-3}
    \multirow{4}{*}{Optimizer} & algorithm & Adam \\
    & learning rates & 1e-3 \\
    & betas & (0.9, 0.98) \\
    & max grad norm & 5.0 \\
    \cmidrule(lr){1-1} \cmidrule(lr){2-2} \cmidrule(lr){3-3}
    \multirow{3}{3cm}{Training} & batch size & 2,048 tokens \\
    & sample-break-mode & none \\
    & epochs & 10 \\
    \cmidrule(lr){1-1} \cmidrule(lr){2-2} \cmidrule(lr){3-3}
    \multirow{3}{3cm}{Inference} & beam size & 100 \\
    & word beam size & 10 \\
    & shift size & 1 \\ \bottomrule
        \end{tabular}
        \subcaption{RNNG.}
        \label{tbl:hyperparam_rnng}
        \vspace{0.2cm}
\end{minipage}

\begin{minipage}[t]{\hsize}
\renewcommand{\arraystretch}{0.5}
    \centering
    \begin{tabular}{p{3cm}p{5cm}p{4.5cm}} \toprule
     \multirow{7}{3cm}{model} & composition & Simple RNN \\
      & recurrence & Simple RNN \\
     & embed\_dim & 64 \\
      & hiden\_size & 64 \\
      & layers & 2 \\
      & dropout & 0.3 \\
      & \#params. & 68K \\
    \cmidrule(lr){1-1} \cmidrule(lr){2-2} \cmidrule(lr){3-3}
    \multirow{4}{*}{Optimizer} & algorithm & Adam \\
    & learning rates & 1e-3 \\
    & betas & (0.9, 0.98) \\
    & max grad norm & 5.0 \\
    \cmidrule(lr){1-1} \cmidrule(lr){2-2} \cmidrule(lr){3-3}
    \multirow{3}{3cm}{Training} & batch size & 2,048 tokens \\
    & sample-break-mode & none \\
    & epochs & 10 \\
    \cmidrule(lr){1-1} \cmidrule(lr){2-2} \cmidrule(lr){3-3}
    \multirow{3}{3cm}{Inference} & beam size & 100 \\
    & word beam size & 10 \\
    & shift size & 1 \\ \bottomrule
        \end{tabular}
        \subcaption{SRNNG.}
        \label{tbl:hyperparam_srnng}
        \vspace{0.2cm}
\end{minipage}

\caption{Hyperparameters of RNNGs}
\label{tbl:hyper_params_rnngs}
\end{table*}

\end{document}